\documentclass[10pt]{article} % For LaTeX2e
\usepackage[accepted]{tmlr}
\usepackage{amsmath,amsfonts,bm}

\def\eqref#1{equation~\ref{#1}}
\def\1{\bm{1}}

\DeclareMathAlphabet{\mathsfit}{\encodingdefault}{\sfdefault}{m}{sl}
\SetMathAlphabet{\mathsfit}{bold}{\encodingdefault}{\sfdefault}{bx}{n}

\usepackage{hyperref}
\usepackage{url}
\usepackage{booktabs}       % professional-quality tables
\usepackage{relsize}
\usepackage{xy}
\usepackage{pifont}
\usepackage{algorithmicx}
\usepackage{algorithm}
\usepackage{algpseudocode}
\xyoption{all}
\usepackage{plain}
\usepackage{xcolor}
\usepackage{amsthm}
\usepackage{graphicx}
\usepackage{makecell}
\usepackage{amsmath,amssymb}
\usepackage[english]{babel}
\newtheorem{theorem}{Theorem}[subsection]

\newtheorem{definition}[theorem]{Definition}
\newtheorem{proposition}[theorem]{Proposition}

\newtheorem{example}[theorem]{Example}
\usepackage{xcolor}

\title{Grothendieck Graph Neural Networks Framework: An Algebraic Platform for Crafting Topology-Aware GNNs}

\author{\name Amirreza Shiralinasab Langari \email amirreza.shiralinasab-langari.1@ens.etsmtl.ca \\
      \addr Department of Electrical Engineering\\
      \'Ecole de Technologie Sup\'erieure (\'ETS), University of Quebec, Montreal, Canada
      \AND
      \name Leila Yeganeh \email leila.yeganeh.1@ens.etsmtl.ca \\
      \addr Department of Electrical Engineering\\
      \'Ecole de Technologie Sup\'erieure (\'ETS), University of Quebec, Montreal, Canada
      \AND
      \name Kim Khoa Nguyen \email kimkhoa.nguyen@etsmtl.ca\\
      \addr Department of Electrical Engineering\\
      \'Ecole de Technologie Sup\'erieure (\'ETS), University of Quebec, Montreal, Canada}

\def\month{04}  % Insert correct month for camera-ready version
\def\year{2026} % Insert correct year for camera-ready version
\def\openreview{\url{https://openreview.net/forum?id=oD3qWXgB4e}} % Insert correct link to OpenReview for camera-ready version

\begin{document}

\maketitle

\begin{abstract}
Graph Neural Networks (GNNs) are almost universally built on a single primitive: the neighborhood. Regardless of architectural variations, message passing ultimately aggregates over neighborhoods, which intrinsically limits expressivity and often yields power no stronger than the Weisfeiler–Lehman (WL) test. In this work, we challenge this primitive. We introduce the Grothendieck Graph Neural Networks (GkGNN) framework, which provides a strict algebraic extension of neighborhoods to \emph{covers}, and in doing so replaces neighborhoods as the fundamental objects of message passing. Neighborhoods and adjacency matrices are recovered as special cases, while covers enable a principled and flexible foundation for defining topology-aware propagation schemes.
GkGNN formalizes covers and systematically translates them into matrices, analogously to how adjacency matrices encode neighborhoods, enabling both theoretical analysis and practical implementation. Within this framework, we introduce the \emph{cover of sieves}, inspired by category theory, which captures rich topological structure. Based on this cover, we design \emph{Sieve Neural Networks} (SNN), a canonical fixed-cover instantiation that generalizes the adjacency matrix. Experiments show that SNN achieves zero observed failures on challenging graph isomorphism benchmarks (SRG, CSL, BREC) and substantially improves topology-aware evaluation via a controlled label-propagation probe. These results establish GkGNN as a principled \emph{foundational framework} for replacing neighborhoods in GNNs.
\end{abstract}

\section{Introduction}

The concept of \emph{neighborhood} plays a central role in most Graph Neural Network (GNN) architectures, serving as the foundation of message passing \citep{gilmer}. This reliance is not arbitrary: neighborhoods provide comprehensive coverage of the graph structure, leveraging the adjacency matrix to facilitate efficient and systematic aggregation of local information. However, this neighborhood-based perspective comes with intrinsic limitations. In particular, many GNNs have expressive power bounded by the WL test \citep{sato}, \citep{gin}, restricting their ability to capture broader topological structures.

To address the limitations of neighborhoods, researchers have proposed alternatives that incorporate richer structural information. One direction uses concepts from algebraic topology, such as simplicial complexes and higher-order faces, to capture interactions beyond pairs of nodes \citep{sc}, \citep{cw}, \citep{tdl}, \citep{topotune}. Another line of work relies on specific patterns or subgraphs (e.g., motifs) to encode characteristic structures as the basis for topologically-aware message passing \citep{gsn}, \citep{tlgnn}. While these methods enrich the local perspective, they often depend on handcrafted definitions or combinatorial choices. In contrast, neighborhoods themselves arise from a precise algebraic definition, suggesting that a systematic algebraic formalism for extending neighborhoods may provide a broader and more principled foundation for message passing itself.

\paragraph{Viewpoint.}
We argue that neighborhoods should no longer be regarded as the primitive objects of message passing. Instead, we propose to replace neighborhoods by \emph{covers}\footnote{The term \emph{cover} here should not be confused with graph-theoretic vertex or edge covers; it refers to a collection of algebraically defined message-passing strategies.}
 as the fundamental units over which message passing operates. Covers strictly generalize neighborhoods: neighborhoods are recovered as a special case, while covers enable algebraically defined alternatives to neighborhood aggregation that go beyond locality. The purpose of this paper is to establish the algebraic and invariance foundations of this replacement. In particular, we introduce the first algebraic formalism for systematically generating and implementing infinitely many message-passing strategies, ranging from standard neighborhood aggregation to covers that encode richer topological structure. Covers may be instantiated as fixed algebraic objects, as in this work, or parameterized and learned in future architectures.
Throughout this paper, replacing neighborhoods refers to replacing them as the primitive abstraction of message passing, not to discarding adjacency-based information: neighborhoods appear as specific elements within a larger algebraic space of composable propagation operators.

Building on this viewpoint, our work introduces an algebraic framework that extends neighborhoods to covers while preserving their structural role. This leads to the \emph{Grothendieck Graph Neural Networks} (GkGNN) framework, an algebraic platform that formalizes covers, studies their invariance properties, and systematically translates them into matrix representations. In this framework, the adjacency matrix appears as a special case, while more expressive covers give rise to new message-passing operators.

Because this work extends the notion of neighborhoods, any existing GNN architecture can, in principle, be reformulated to operate on covers instead of neighborhoods. Accordingly, our focus is not on proposing a new learning algorithm, but on establishing the algebraic and invariance foundations that make such a replacement possible. At the core of GkGNN is a duality between graphs and algebraic structures, showing that graphs can be characterized by monoids generated from their covers.

Our main contributions are as follows.
\begin{itemize}
\item \textbf{Replacing neighborhoods by covers.}
We introduce the concept of \emph{covers} for graphs as a strict algebraic extension of neighborhoods. This generalization provides a principled and flexible foundation for representing graph structure.

\item \textbf{The GkGNN framework.}
We develop the \emph{Grothendieck Graph Neural Networks} (GkGNN) framework, which creates, refines, and transforms covers into their matrix forms, recovering the adjacency matrix as a special case. GkGNN offers a systematic way to design new message-passing strategies.

\item \textbf{Sieve Neural Networks (SNN).}
As a concrete instantiation of GkGNN, we introduce \emph{Sieve Neural Networks} (SNN) as a canonical example showing how expressive algebraic covers can be transformed into message-passing operators.

\item \textbf{Topology-aware evaluation.}
In addition to evaluating SNN on graph isomorphism tasks, we design a topology-encoding benchmark based on a special case of Label Propagation (LP)~\citep{lp1,lp2} with one propagation step $(\alpha = 1)$. Here LP is used not as a learning algorithm but as a controlled probe to directly compare covers with neighborhoods. On citation networks and ogbn-arxiv, sieve covers consistently and substantially outperform the neighborhood cover, highlighting the advantage of covers in capturing topological structure in large graphs.
\end{itemize}

The proposed extension of the neighborhood concept in this paper relies on
introducing four monoids (see Appendix~\ref{defapp} for formal definition of a monoid and monoidal homomorphism),
$(\mathsf{Mod}(G),\bullet)\subseteq(\mathsf{SMult}(G),\bullet)$ and
$(\mathsf{Mom}(G),\circ)\subseteq(\mathsf{Mat}_{n}(\mathbb{R}),\circ)$.
In this setting, neighborhoods are elements of $\mathsf{Mod}(G)$, and a
monoidal homomorphism
$\mathsf{Tr}:\mathsf{Mod}(G)\rightarrow\mathsf{Mom}(G)$ translates elements
of $\mathsf{Mod}(G)$ into matrices.

This organization allows us to treat graphs as algebraic spaces, in a manner
analogous to familiar examples such as $(\mathbb{N},\times)$, where natural
numbers are composed by multiplication, or
$(\mathsf{Mat}_{n}(\mathbb{R}),\cdot)$, where matrices are composed by matrix
multiplication. In the same spirit, the monoid operation $\bullet$ enables
algebraic compositions of graph-derived objects, for instance
$N(u)\bullet N(v)$.

As proved in Theorem~\ref{basis}, directed edges generate $\mathsf{Mod}(G)$.
Intuitively, directed edges act as elementary building blocks: by composing
edges as $e_1\bullet e_2\bullet\cdots\bullet e_k$, one can generate both
neighborhoods and elements beyond neighborhoods, in an informal sense
analogous to how the monoid $(\mathbb{N},\times)$ is generated by prime numbers
under multiplication.

The map $\mathsf{Tr}$ plays an essential role in the GkGNN framework: it is not
merely a function, but a monoidal homomorphism. For example, the determinant
defines a monoidal homomorphism from
$(\mathsf{Mat}_{n}(\mathbb{R}),\cdot)$ to $(\mathbb{R},\times)$, since
$\det(A\cdot B)=\det(A)\times\det(B)$.
Since $\mathsf{Tr}$ is a monoidal homomorphism, the translation of a composed
element $e_1\bullet e_2\bullet\cdots\bullet e_k\in\mathsf{Mod}(G)$ can be
computed compositionally as
\(
\mathsf{Tr}(e_1\bullet e_2\bullet\cdots\bullet e_k)
=
\mathsf{Tr}(e_1)\circ \mathsf{Tr}(e_2)\circ\cdots\circ \mathsf{Tr}(e_k),
\)
which preserves the algebraic structure of compositions when passing from
graph-based objects to matrices.

\section{Related work}\label{section 2}
Many classical GNN architectures can be unified under the neighborhood-based \emph{Message Passing Neural Network} (MPNN) paradigm~\citep{gilmer}. A large body of work seeks to move beyond strict 1-hop neighborhoods by altering the graph on which messages are passed or by enriching the operators/features used for aggregation.

\paragraph{Passing messages on derived graphs.}
One line of work replaces the original graph with a derived graph and then applies MPNN. For example, \cite{dirmpnn} constructs the \emph{directed line graph}, whose nodes correspond to directed edges of the original graph and where two nodes are adjacent if the underlying edges share an endpoint; message passing is then performed on this derived graph. In \cite{tlgnn}, each graph is mapped to a \emph{topology-level} summary graph built from subgraphs; message passing runs jointly on the original graph and its summary, making the propagation explicitly topology-aware.

\paragraph{Substructure and kernel based encodings.}
Another direction injects information about motifs or subgraphs. Graph Substructure Networks (GSNs)~\citep{gsn} enrich node/edge features with positions within selected patterns, integrating substructure signals into message passing. KerGNNs~\citep{kergnn} use small graphs as filters—via graph kernels such as random-walk kernels—applied to node-centered subgraphs; replacing the raw neighborhood with a filtered subgraph can increase expressivity over vanilla MPNNs.

\paragraph{Contextual and multi-hop neighborhoods.}
Contextualization beyond the immediate neighborhood is also common. ID-GNN~\citep{idgnn} attends to occurrences of a node within its ego network, effectively differentiating its roles across contexts. Extensions to $k$-hop neighborhoods~\citep{khop} aggregate information from larger receptive fields; KP-GNN further selects $k$-hop neighbors via shortest-path or random-walk kernels, yielding a framework that can surpass standard MPNNs.

\paragraph{Local topology operators and stochastic perturbations.}
In \cite{vig}, each node is associated with a \emph{local context} matrix intended to capture surrounding topology; these contexts replace raw features during message passing and have been shown effective on topology-sensitive tasks (e.g., cycle detection), outperforming MPNNs in those settings. The approach in \cite{drop} applies message passing to randomly thinned graphs obtained by deleting each node with small probability and aggregates the outcomes, preserving much of the original topology while introducing beneficial stochasticity.

\paragraph{Topological deep learning.}
Tools from algebraic topology provide higher-order generalizations of graphs that encode multi-level interactions. Works based on simplicial and CW complexes~\citep{sc,cw} replace node–edge neighborhoods with higher-dimensional cells and associated incidence structures, yielding message-passing schemes that explicitly reason over topology beyond pairwise relations.

\section{Covers and their matrix interpretations}
In this section we develop the notion of \emph{covers} for graphs and show how to interpret them as matrices, laying the groundwork for the GkGNN framework.
We begin by assigning to each directed subgraph its matrix representation,
establishing a bijection between directed subgraphs and their associated
matrices. We then introduce two monoids: $\mathsf{Mod}(G)$, generated by
directed subgraphs, and $\mathsf{Mom}(G)$, generated by
their matrix representations.
This allows us to extend the representation map to a monoidal homomorphism
\[
  \mathsf{Tr}:\ \mathsf{Mod}(G)\ \longrightarrow\ \mathsf{Mom}(G).
\]
We prove that $\mathsf{Tr}$ is invariant under graph isomorphisms (via
Change-of-Order mappings) and provides an algebraic description of a graph
that is unique up to isomorphism. These results form the theoretical
foundation of the GkGNN framework.

\subsection{Matrix representations of directed subgraphs}

We consider undirected graphs $G=(V,E)$ whose node set $V$ is equipped with a fixed ordering. Our first step is to formalize \emph{directed subgraphs} of $G$ and to define their matrix representations.

\begin{definition}\label{def of path and dirsubgraph}

(1) A \textbf{path} $p$ from node $v_{p_1}$ to node $v_{p_m}$ is an ordered sequence
    \[
      v_{p_1},\, e_{p_1},\, v_{p_2},\, e_{p_2},\, \dots,\, v_{p_{m-1}},\, e_{p_{m-1}},\, v_{p_m},
    \]
    where each $e_{p_i}$ connects $v_{p_i}$ and $v_{p_{i+1}}$.
    
(2) A \textbf{directed subgraph} $D$ of $G$ is a connected, acyclic
subgraph in which each edge is assigned a \emph{direction}.
\end{definition}

\paragraph{Neighborhoods as a special case.}
A neighborhood can be seen as a directed subgraph obtained by orienting
all incident edges \emph{into} a fixed node (see Figure~\ref{sink}). In the
adjacency matrix, each column corresponds to such a neighborhood; isolating the
neighborhood of a node amounts to zeroing out the other columns.

\paragraph{Matrix representation of a directed subgraph.}
We extend the neighborhood-as-column view to arbitrary directed subgraphs by
encoding \emph{direction-respecting reachability}.

\begin{definition}\label{def_of_mat_rep}
Let $D$ be a directed subgraph of $G=(V,E)$.
\begin{enumerate}\itemsep4pt
    \item Define a relation $\le_D$ on $V$ by $v_i \le_D v_j$ iff there exists a path
    in $D$ that \emph{respects edge directions} and starts at $v_i$ and ends at $v_j$.
    \item The \emph{matrix representation} of $D$ is the $|V|\times |V|$ matrix
    $M_D$ with $(M_D)_{ij}=1$ if $v_i \le_D v_j$ and $(M_D)_{ij}=0$ otherwise.
\end{enumerate}
\end{definition}
The requirement that paths respect directions is essential: all walks counted by $M_D$ must follow the edge orientations in $D$. Intuitively, a directed subgraph specifies an \emph{allowable message-flow pattern}; its matrix $M_D$ realizes this pattern. In this sense, matrices from directed subgraphs serve as structured alternatives to the standard adjacency matrix used in neighborhood-based message passing.

\begin{figure}[ht]
    \centering
    \includegraphics[scale=0.4]{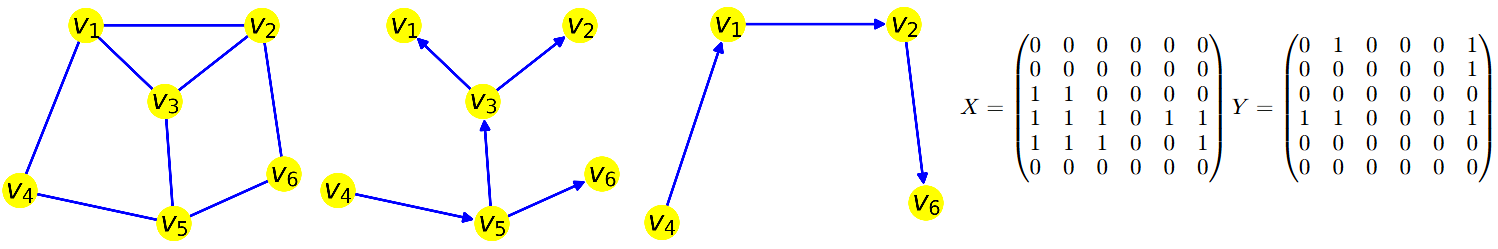}
    \caption{\textbf{Directed subgraphs and their matrices.} Two directed subgraphs
    of a graph $G$ (left) are shown: $\hat{D}$ (middle) and $\bar{D}$ (right).
    Their matrix representations are $X$ and $Y$, respectively. Each directed
    subgraph encodes a distinct strategy for propagating information; the
    matrices $X$ and $Y$ make these strategies directly usable in message passing.}
    \label{directed subgraph}
\end{figure}

\paragraph{Representation map.}
Definition~\ref{def_of_mat_rep} induces a map from directed subgraphs to matrices:
\[
  \mathsf{Rep}:\ \mathsf{DirSub}(G)\ \longrightarrow\ \mathsf{MatRep}(G),
\]
where $\mathsf{DirSub}(G)$ is the set of directed subgraphs of $G$ and
$\mathsf{MatRep}(G)$ is the image (subset) of $\mathsf{Mat}_{|V|}(\mathbb{R})$
consisting of matrices that arise from directed subgraphs via
Definition~\ref{def_of_mat_rep}.

\begin{theorem}\label{phi}
The map $\mathsf{Rep}$ is an isomorphism between $\mathsf{DirSub}(G)$ and
$\mathsf{MatRep}(G)$. In particular, each directed subgraph is uniquely determined by its matrix representation, and conversely every matrix in $\mathsf{MatRep}(G)$ corresponds to a unique directed subgraph.
\end{theorem}

\subsection{Defining covers for graphs: an algebraic platform}

While we can cover a graph $G$ by picking elements from $\mathsf{DirSub}(G)$ and
map each to a matrix via $\mathsf{Rep}$, this space is limited: $\mathsf{DirSub}(G)$
is relatively small and its elements do not combine well. For instance, the union
of the directed subgraphs $\hat D$ and $\bar D$ in Figure~\ref{directed subgraph}
is \emph{not} a directed subgraph (multiple directed paths appear between some
node pairs). Hence no matrix image exists for such a combination, which prevents
its direct use in a message-passing scheme. This makes it hard to design diverse,
meaningful strategies using only $\mathsf{DirSub}(G)$.

\paragraph{Step 1: enlarge the space via a multigraph monoid.}
To combine directed subgraphs systematically, we first endow them with an algebraic operation. A natural choice is to define $C \bigoplus D$ for $C,D \in \mathsf{DirSub}(G)$ as the \emph{directed multigraph} obtained by taking the union of their node sets and the disjoint union of their directed edge sets. Let
{\small\[
  \mathsf{Mult}(G)=\Big\{\;\bigoplus_{i=1}^{k} D_i \;:\;
  k\ge 1,\; D_i\in\mathsf{DirSub}(G)\;\Big\}.
\]}
Then $(\mathsf{Mult}(G),\bigoplus)$ is a commutative monoid.
This construction enlarges the object set, but it treats edges as independent
and is largely \emph{insensitive to path structure}: paths in $C\bigoplus D$
are simply concatenations of available directed edges, with little inheritance
from the path sets of $C$ and $D$. As a result, $\bigoplus$ alone provides
limited control for crafting new message-passing strategies.

\paragraph{Step 2: add path sensitivity via a non-commutative monoid.}
To encode which \emph{paths} are allowed, not only which edges exist, we augment multigraphs with explicit path sets. Define
{\small\[
  \mathsf{SMult}(G)
  \;=\;
  \big\{\, (M,S) \;:\; M\in \mathsf{Mult}(G),\; S\subseteq \mathsf{Paths}(M)\,\big\},
\]}
and define a binary operation $\bullet$ by
\(
  (M,S)\bullet (N,T)
  \;=\;
  \big(M\bigoplus N,\; S \star T\big),
\)
where $S\star T$ is the union of the sets $S$, $T$, and the collection of paths constructed by the composition of paths in $S$ followed by paths in $T$. Since edges of $M\bigoplus N$ come from a disjoint union, paths in $S$ and $T$ remain disjoint subsets of $\mathsf{Paths}(M\bigoplus N)$.

\begin{theorem}\label{smult}
$(\mathsf{SMult}(G),\bullet)$ is a non-commutative monoid.
\end{theorem}

Non-commutativity comes from the order of path composition inside $\star$:
if $(M,S)$ and $(N,T)$ have composable paths, then generally
$(M,S)\bullet(N,T) \neq (N,T)\bullet(M,S)$; if they have no composable paths, $\star$ reduces to set union.

\begin{example}\label{noncommutativity}
Let $d:u\!\to\! v$ and $e:v\!\to\! w$ be directed edges.
Then $d\bullet e$ and $e\bullet d$ share the same edge multiset
($d\bigoplus e$), but differ in allowed paths: $d\bullet e$ contains a path
from $u$ to $w$ (via $d$ then $e$), whereas $e\bullet d$ does not.
Thus the order of composition matters.
\end{example}

\paragraph{Step 3: a representable submonoid for covers.}
Elements $(M,S)\in\mathsf{SMult}(G)$ can be read as \emph{message-passing
strategies}: $M$ fixes the directed multigraph over which messages may travel, and $S$ specifies which directed paths are allowed. However, not every such pair admits a matrix representation compatible with an extension of $\mathsf{Rep}$. To retain implementability, we restrict to the submonoid generated by genuine directed subgraphs, embedded via $D \mapsto \big(D,\mathsf{Paths}(D)\big)$.

\begin{definition}
For a graph $G$, the \emph{monoid of directed subgraphs} is the submonoid
$\mathsf{Mod}(G)\subseteq \mathsf{SMult}(G)$ generated by $\mathsf{DirSub}(G)$.
\end{definition}
Hence each $(M,S)\in\mathsf{Mod}(G)$ can be written as
{\small\[
  (M,S) \;=\; D_1 \bullet \cdots \bullet D_k,\qquad
  M \;=\; \bigoplus_{i=1}^{k} D_i,\qquad
  S \;=\; \mathsf{Paths}(D_1)\star \cdots \star \mathsf{Paths}(D_k).
\]}
In the next subsection we show that all elements of $\mathsf{Mod}(G)$ admit
matrix representations via an extension of $\mathsf{Rep}$, which makes them
suitable for implementation.

\begin{definition}\label{def_cover}
A \emph{cover} of $G$ is a finite collection of elements of $\mathsf{Mod}(G)$.
\end{definition}

A cover thus specifies a family of message-passing strategies: each element
constrains allowed paths locally, and the collection provides a flexible,
task-dependent view of the graph (we do not require covering all nodes/edges).
Because $\mathsf{Mod}(G)$ is infinite and $\bullet$ is non-commutative, this
space is expressive enough to encode a wide range of perspectives on how
information should flow. Moreover, the following result shows that directed
edges suffice as generators, so complex strategies can be built compositionally.

\begin{theorem}\label{basis}
Directed edges generate $\mathsf{Mod}(G)$.
\end{theorem}

\subsection{From covers to matrices: an algebraic perspective}

Our goal in this section is to extend the representation map $\mathsf{Rep}$ from directed subgraphs to arbitrary elements of $\mathsf{Mod}(G)$, so that a \emph{cover} can be transformed into a \emph{collection of matrices}. This requires a matrix-side operation that mirrors the path-composition on $\mathsf{Mod}(G)$.

\paragraph{A monoid on matrices.}
Let $\mathsf{Mat}_{n}(\mathbb{R})$ be the set of $n \times n$ real matrices. Define a binary operation
{\small\[
  A \circ B \;:=\; A \;+\; B \;+\; AB .
\]}
\begin{theorem}\label{matmon}
$(\mathsf{Mat}_{n}(\mathbb{R}), \circ)$ is a monoid.
\end{theorem}

We now take the submonoid \emph{generated} by $\mathsf{MatRep}(G)$ inside
$(\mathsf{Mat}_{|V|}(\mathbb{R}),\circ)$.

\begin{definition}
The \emph{monoid of matrix representations} of $G$ is the submonoid
$(\mathsf{Mom}(G),\circ)$ of $(\mathsf{Mat}_{|V|}(\mathbb{R}),\circ)$
generated by $\mathsf{MatRep}(G)$.
\end{definition}

\paragraph{Extending $\mathsf{Rep}$ to covers.}
Elements of $\mathsf{Mod}(G)$ are built by composing directed subgraphs with
$\bullet$. The map below sends such compositions to matrices by replacing each directed subgraph $D_i$ with its matrix $\mathsf{Rep}(D_i)$ and each $\bullet$ with $\circ$.

\begin{theorem}\label{monoidal surjection}
The mapping $\mathsf{Tr}:\mathsf{Mod}(G)\longrightarrow \mathsf{Mom}(G)$,
{\small\[
  (M,S)=D_1\bullet D_2\bullet\cdots \bullet D_k \longmapsto
  A \;=\; A_1\circ A_2\circ\cdots\circ A_k,
\]}
where $D_i\in\mathsf{DirSub}(G)$ and $A_i=\mathsf{Rep}(D_i)$, is a surjective monoidal homomorphism.
\end{theorem}

\paragraph{Interpretation (path counting).}
In the proof of Theorem~\ref{monoidal surjection} one sees that $\mathsf{Tr}$ acts as a \emph{path counter}: for $(M,S)\in\mathsf{Mod}(G)$, the $(i,j)$ entry of $\mathsf{Tr}(M,S)$ equals the number of paths in $S$ from $v_i$ to $v_j$. Thus, covers become \emph{collections of matrices} in the same way that neighborhoods give rise to the adjacency matrix, but now with explicit control over composed paths. Although we were not able to show that $\mathsf{Tr}$ is an isomorphism, it extends an isomorphism on $\mathsf{DirSub}(G)$ and will be shown in the next subsection to characterize graphs up to isomorphism, supporting
its use as a faithful matrix transformation of covers.

Arising from Theorem~\ref{monoidal surjection}, the following commutative diagram
illustrates how $\mathsf{Tr}$ extends the representation map $\mathsf{Rep}$:
\[
\xymatrix{
\mathsf{DirSub}(G)\ar[rr]^{\mathsf{Rep}}\ar@{^{(}->}[d]
&&
\mathsf{MatRep}(G)\ar@{^{(}->}[d]\\
\mathsf{Mod}(G)\ar[rr]_{\mathsf{Tr}}
&&
\mathsf{Mom}(G)
}
\]
Since
$\mathsf{Rep}:\mathsf{DirSub}(G)\to \mathsf{MatRep}(G)$ is an isomorphism,
$\mathsf{Tr}$ is constructed to agree with $\mathsf{Rep}$ on directed subgraphs:
under the canonical embedding $D\mapsto (D,\mathrm{Paths}(D))\in\mathsf{Mod}(G)$,
we have $\mathsf{Tr}(D)=\mathsf{Rep}(D)$. Thus $\mathsf{Tr}$ extends an isomorphism
on $\mathsf{DirSub}(G)$ while enlarging the space to $\mathsf{Mod}(G)$.

The non-commutativity of $\bullet$ (and correspondingly of $\circ$) is essential for
this extension to remain meaningful beyond $\mathsf{DirSub}(G)$, because ordered
compositions encode order of directed edges in paths. To see this, let $D\in\mathsf{DirSub}(G)$
be the path
\[
\xymatrix{u \ar[r]^d & v \ar[r]^e & w \ar[r]^f & z}.
\]
In $\mathsf{Mod}(G)$, the same object can be written using edge generators as
$D = d\bullet e\bullet f=(M,S)$, where
\[
S=\{d,e,f,de,ef,def\},
\]
and $de$, $ef$, $def$ denote successive compositions of edges (subpaths of $D$).
By monoidality,
\[
\mathsf{Tr}(D)=\mathsf{Tr}(d\bullet e\bullet f)
=\mathsf{Tr}(d)\circ\mathsf{Tr}(e)\circ\mathsf{Tr}(f),
\]
and $\mathsf{Tr}$ records exactly the paths in $S=\mathrm{Paths}(D)$ (via
its path-counting interpretation), hence $\mathsf{Tr}(D)=\mathsf{Rep}(D)$ on
$\mathsf{DirSub}(G)$. If $\bullet$ were commutative, ordered compositions such as
$d\bullet e$ and $e\bullet d$ would coincide, collapsing distinct path constraints
and destroying the order-sensitive information that $\mathsf{Tr}$ is meant to carry
when extending $\mathsf{Rep}$ beyond single directed subgraphs.

\begin{example}
    In Figure~\ref{directed subgraph}, let $\hat D$ and $\bar D$ be two directed subgraphs of $G$ with matrix representations $X$ and $Y$. Using $\bullet$, we may form new strategies $\hat D \bullet \bar D$ and $\bar D \bullet \hat D$. Their matrix transforms are obtained as follows:
{\small\[
  \mathsf{Tr}(\hat D \bullet \bar D)
  \;=\; \mathsf{Tr}(\hat D)\circ \mathsf{Tr}(\bar D)
  \;=\; X \circ Y,
  \qquad
  \mathsf{Tr}(\bar D \bullet \hat D)
  \;=\; \mathsf{Tr}(\bar D)\circ \mathsf{Tr}(\hat D)
  \;=\; Y \circ X.
\]}
Since $\circ$ is generally non-commutative in this context, the two results
differ, reflecting the order-sensitivity of path composition in the underlying cover construction.
\end{example}

\section{Grothendieck Graph Neural Networks Framework}

\subsection{Algebraic foundations of graphs}
We have introduced two monoids associated with a graph and a monoidal homomorphism
between them. We now ask: \emph{to what extent do these algebraic objects describe
the underlying graph?} To address this, we first formalize how reordering node
indices acts on the matrix space and verify that this action is compatible with
our monoidal structures.

\paragraph{Change-of-Order mappings.}
A matrix $A \in \mathsf{Mat}_{n}(\mathbb{R})$ represents a linear map
$\mathbb{R}^{n} \to \mathbb{R}^{n}$ with respect to the standard basis.
If we reorder the standard basis of $\mathbb{R}^{n}$ (equivalently, relabel the
coordinates), the matrix representation of the same linear map is obtained by
reindexing the rows and columns of $A$. Any linear isomorphism
$f:\mathsf{Mat}_{n}(\mathbb{R}) \to \mathsf{Mat}_{n}(\mathbb{R})$ arising in this
way is called a \textbf{Change-of-Order mapping} (see Example~\ref{exco}).
Intuitively, this captures the effect of node relabeling at the matrix level.

\begin{proposition}\label{changeindices}
Suppose $f:\mathsf{Mat}_{n}(\mathbb{R}) \rightarrow \mathsf{Mat}_{n}(\mathbb{R})$
is a Change-of-Order mapping. Then $f$ preserves the standard algebraic operations
on matrices: it is compatible with monoidal operation \(\circ\), with matrix multiplication,
and with element-wise (Hadamard) multiplication.
\end{proposition}

\paragraph{Relating graph isomorphisms and algebra.}
A graph isomorphism $f: G \to H$ is a bijection on nodes that preserves edges,
and therefore corresponds to a reordering of node indices. Hence it induces a
Change-of-Order mapping $\mathsf{CO}(f): \mathsf{Mat}_{|V_G|}(\mathbb{R}) \to
\mathsf{Mat}_{|V_H|}(\mathbb{R})$. The next result shows that this relabeling is
compatible with our monoidal constructions and with the translation to matrices.

\begin{theorem}\label{iso_graph-induce}
Every graph isomorphism $f: G \rightarrow H$ induces monoidal isomorphisms
$\mathsf{Mod}(f): \mathsf{Mod}(G) \rightarrow \mathsf{Mod}(H)$ and
$\mathsf{Mom}(f): \mathsf{Mom}(G) \rightarrow \mathsf{Mom}(H)$ such that the
following diagram commutes, where $\iota$ denotes the inclusions:
\small{
\begin{equation} \label{graph iso implies com}
    \xymatrix{
        \mathsf{Mod}(G) \ar[r]^{\mathsf{Tr}_G} \ar[d]_{\mathsf{Mod}(f)} &
        \mathsf{Mom}(G) \ar[d]_{\mathsf{Mom}(f)} \ar@{^{(}->}[r]^{\iota} &
        \mathsf{Mat}_{|V_G|}(\mathbb{R}) \ar[d]^{\mathsf{CO}(f)} \\
        \mathsf{Mod}(H) \ar[r]_{\mathsf{Tr}_H} &
        \mathsf{Mom}(H) \ar@{^{(}->}[r]_{\iota} &
        \mathsf{Mat}_{|V_H|}(\mathbb{R})
    }
\end{equation}
}
\end{theorem}

\paragraph{A converse direction.}
We also have a partial converse: if a Change-of-Order mapping identifies the
matrix-level monoids of two graphs, then the graphs are isomorphic.

\begin{theorem}\label{induce_iso_graph}
Let $G$ and $H$ be graphs with $|V_G| = |V_H| = n$, and let
$f:\mathsf{Mat}_{n}(\mathbb{R}) \rightarrow \mathsf{Mat}_{n}(\mathbb{R})$ be a
Change-of-Order mapping. If the restriction of $f$ to $\mathsf{Mom}(G)$ is an
isomorphism onto $\mathsf{Mom}(H)$, then $G$ and $H$ are isomorphic.
\end{theorem}
These theorems establish a duality between graphs and monoids, in which elements of \(\mathsf{Mod}(G)\) encode structural relations intrinsic to the graph. Importantly, this duality does not originate from neighborhoods alone, since neighborhoods are known to be insufficient to characterize graphs, but from richer elements of \(\mathsf{Mod}(G)\) that capture more global, structure-aware information.

\subsection{Definition of the GkGNN framework}\label{subsection 3.5}
Theorems~\ref{iso_graph-induce} and~\ref{induce_iso_graph} establish the key invariance principle underlying our construction. In Diagram~\ref{graph iso implies com}, an isomorphism between graphs $G$ and $H$ corresponds to the vertical homomorphisms being isomorphisms; equivalently, a relabeling of nodes in a graph induces isomorphic transformations both on a cover and on its matrix transformation.
Consequently, the \emph{horizontal} arrows in the diagram provide an algebraic description that is unique up to graph isomorphism. Leveraging this observation, we formalize the \emph{Grothendieck Graph Neural Networks (GkGNN)} framework as the following algebraic pipeline:
\begin{definition}\label{ggnn_framework}
    For a graph $G=(V,E)$, the GkGNN framework is the composition
    {\small\begin{equation}\label{GGNNF}
      \xymatrix{\mathsf{Mod}(G)\ar[r]^{\mathsf{Tr}}&\mathsf{Mom}(G)\ar@{^{(}->}[r]^{\iota}&\mathsf{Mat}_{|V|}(\mathbb{R})}   \end{equation}}
\end{definition}
%%%%%%%%%%%%%%%%%%%%%
\paragraph{How GkGNN is used.}
The framework exposes three actions:
(i) \emph{choose} a cover in $\mathsf{Mod}(G)$;
(ii) \emph{translate} it to matrices via $\mathsf{Tr}$;
(iii) optionally \emph{enrich} the resulting collection inside
$\mathsf{Mat}_{|V|}(\mathbb{R})$ using the allowed operations from
Proposition~\ref{changeindices}. See Appendix~\ref{expofggnn} for more details.

\paragraph{Neighborhoods as a special case.}
The framework recovers standard message passing from neighborhoods:

\begin{theorem}\label{cover of neighborhoods}
The collection of neighborhoods forms a cover in $\mathsf{Mod}(G)$ and, under
$\mathsf{Tr}$, maps to the adjacency matrix in $\mathsf{Mat}_{|V|}(\mathbb{R})$.
\end{theorem}

\paragraph{Remarks.}
(i) A graph can be characterized by the submonoid generated by its neighborhoods; see Appendix~\ref{submonoid_of_neigh}. (ii) We compare GkGNN with higher-order GNN families in Appendix~\ref{comparison_ggnn_higher_orders}.

\section{Sieve Neural Networks: a model within the GkGNN framework}\label{section_snn}

We emphasize that Sieve Neural Networks ($\mathsf{SNN}$) is not the goal of GkGNN, but a canonical example demonstrating how expressive algebraic elements of \(\mathsf{Mod}(G)\) can be identified and transformed into concrete message-passing operators.
$\mathsf{SNN}$ is deliberately fixed and non-learned, in order to isolate the representational effect of covers and to exemplify how elements beyond neighborhoods can be used in practice.

Inspired by \emph{sieves} in category theory \citep{macmo}, we instantiate the GkGNN
framework with a concrete cover that yields the \emph{Sieve Neural Network}. For each node $v$, we construct a family of outward-expanding
substructures, formalized as elements of $\mathsf{Mod}(G)$, which together form a \emph{sieve-inspired cover} of the graph. This cover generalizes the standard
neighborhood view by admitting multiple direction-respecting pathways for
information exchange. Via the matrix map $\mathsf{Tr}$, the cover is translated
into operators used for message passing, exposing richer topological relationships
than adjacency-based aggregation while remaining permutation-consistent. In what
follows, we define the sieve and cosieve elements, assemble the cover, and derive
the $\mathsf{SNN}$ architecture from their matrix transformations.

\subsection{Cover of sieves for graphs}

\begin{figure}[t]
    \centering
    \includegraphics[scale=0.3]{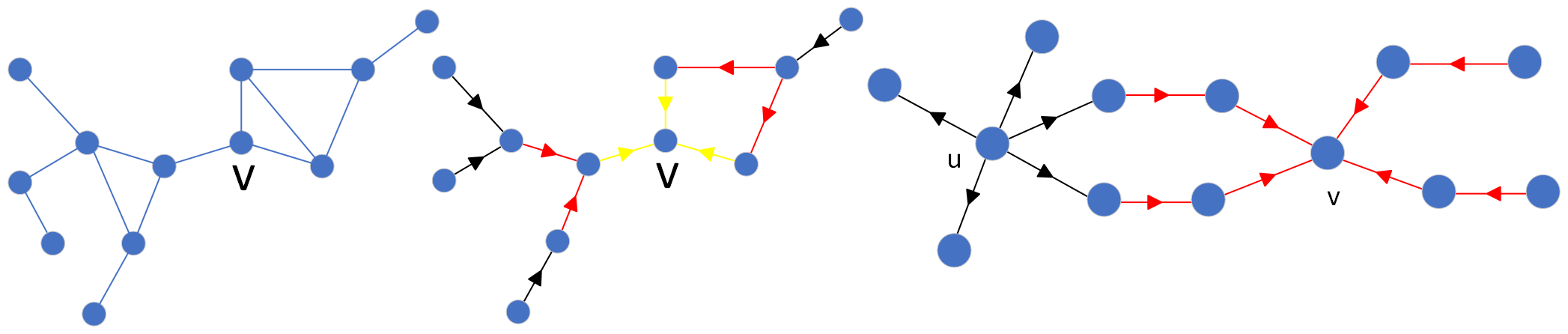}
    \caption{\textbf{Sieve construction.} Left: a graph $G$. Middle:
    $\mathsf{Sieve}(v,3)=D_3(v)\bullet D_2(v)\bullet D_1(v)$ around $v$; yellow, red,
    and black edges indicate $D_1(v)$, $D_2(v)$, and $D_3(v)$, respectively. Right: a graph
    $H$; the element $\mathsf{CoSieve}(u,1)\bullet \mathsf{Sieve}(v,2)\in \mathsf{Mod}(H)$
    specifies allowed interactions between $u$ and $v$ in $\mathsf{SNN}(\alpha,(1,2))$.}
    \label{sieve}
\end{figure}

\paragraph{Constructing sieve elements in $\mathsf{Mod}(G)$.}
Fix a node $v$ and build breadth-first ``layers'' around $v$:
{\small\[
N_0(v)=\{v\},\qquad N_1(v)=N(v),\qquad
N_k(v)=\Big(\!\!\bigcup_{u\in N_{k-1}(v)}\!\! N(u)\Big)\setminus \bigcup_{i=0}^{k-1} N_i(v)\quad (k\ge 2).
\]}
For each $k\ge 1$, orient all edges \emph{toward} $v$ across consecutive layers and collect them as
\[
M_k(v)=\{\, w\!\to\! u\;:\; wu\in E,\; w\in N_k(v),\; u\in N_{k-1}(v)\,\},
\qquad M_0(v)=\varnothing.
\]
Edges in the same $M_k(v)$ are pairwise non-composable (each goes from layer $k$ to $k{-}1$), so the order in which they are combined is irrelevant. Define
\[
D_k(v)\;:=\;\mathop{\bullet}\limits_{e\in M_k(v)} e
\]
and assemble the depth-$k$ \emph{sieve element}
\[
\mathsf{Sieve}(v,k)\;:=\; D_k(v)\bullet D_{k-1}(v)\bullet\cdots\bullet D_1(v)\bullet D_0(v),
\]
where $D_0(v)$ is the identity of $\mathsf{Mod}(G)$; see Figure~\ref{sieve}.
Since the layers eventually empty, there exists $k_0$ with $N_{k_0+1}(v)=\varnothing$, hence
$\mathsf{Sieve}(v,k)$ stabilizes for $k\ge k_0$. We denote this saturated element by $\mathsf{Sieve}(v,-1):=\mathsf{Sieve}(v,k_0)$.
To construct the \emph{opposite} sieve, reverse the directions in each $M_k(v)$:
let $M_k^{\mathrm{op}}(v)$ be $M_k(v)$ with all edges reversed and set
\[
D_k^{\mathrm{op}}(v)\;:=\;\mathop{\bullet}\limits_{e\in M_k^{\mathrm{op}}(v)} e,
\qquad
\mathsf{CoSieve}(v,\ell)\;:=\; D_0^{\mathrm{op}}(v)\bullet D_1^{\mathrm{op}}(v)\bullet\cdots\bullet D_\ell^{\mathrm{op}}(v).
\]

\paragraph{The cover of sieves.}
For every node $v$ in $G$, collect all depth-truncated and saturated sieve and co-sieve elements:
{\small\[
    \mathsf{Sieve}(v,0),\; \mathsf{Sieve}(v,1),\;\dots,\; \mathsf{Sieve}(v,-1)
\quad\text{and}\quad
\mathsf{CoSieve}(v,0),\; \mathsf{CoSieve}(v,1),\;\dots,\; \mathsf{CoSieve}(v,-1).
\]}

The \emph{cover of sieves} is the finite collection containing these elements for all nodes $v$ in $G$.

\paragraph{Matrix interpretation of the cover of sieves.}
Apply the monoidal homomorphism $\mathsf{Tr}$ to obtain matrices:
{\small\[
\mathsf{Image}(v,k):=\mathsf{Tr}\big(\mathsf{Sieve}(v,k)\big),
\qquad
\mathsf{CoImage}(v,\ell):=\mathsf{Tr}\big(\mathsf{CoSieve}(v,\ell)\big).
\]}
Since $\mathsf{Tr}$ is monoidal,
{\small\begin{equation}
\label{description of image}
\begin{aligned}
\mathsf{Image}(v,k)
&= \mathsf{Tr}\big(D_k(v)\bullet D_{k-1}(v)\bullet\cdots\bullet D_0(v)\big)\\
&= \mathsf{Tr}\big(D_k(v)\big)\circ \mathsf{Tr}\big(D_{k-1}(v)\big)\circ\cdots\circ \mathsf{Tr}\big(D_0(v)\big).
\end{aligned}
\end{equation}}
Within a fixed layer $i$, edges in $M_i(v)$ are not composable, so for distinct
$e,c\in M_i(v)$ we have $\mathsf{Tr}(e)\,\mathsf{Tr}(c)=\mathsf{Tr}(c)\,\mathsf{Tr}(e)=0$.
Hence, by Theorem~\ref{representation},
{\small\[
\mathsf{Tr}\big(D_i(v)\big)
= \mathsf{Tr}\!\Big(\mathop{\bullet}\limits_{e\in M_i(v)} e\Big)
= \mathop{\circ}\limits_{e\in M_i(v)} \mathsf{Tr}(e)
= \sum_{e\in M_i(v)} \mathsf{Tr}(e),
\]}
i.e., $\mathsf{Tr}\big(D_i(v)\big)$ is obtained from the adjacency matrix by keeping only
entries corresponding to directed edges in $M_i(v)$. Moreover,
$\mathsf{CoImage}(v,\ell)=\mathsf{Image}(v,\ell)^\top$ (transpose), so it suffices to compute one of them.
Algorithms for these computations are given in Appendix~\ref{algorithm}. In addition, Appendix~\ref{submonoid_of_sieves} shows that the submonoid generated by this cover determines the graph.

\paragraph{Invariance.}
The cover of sieves is stable under graph isomorphisms, and the induced matrices transform
accordingly.

\begin{theorem}\label{coversieveinvariant}
If $f:G\rightarrow H$ is a graph isomorphism, then
$\mathsf{Mod}(f)\big(\mathsf{Sieve}(v,k)\big)=\mathsf{Sieve}\big(f(v),k\big)$ and
$\mathsf{Mom}(f)\big(\mathsf{Image}(v,k)\big)=\mathsf{Image}\big(f(v),k\big)$.
\end{theorem}

%%%%%%%%%%%%%%%%%%%%%%%%%%%%%%%%%%%%%%%%%%5
\subsection{Design and construction of the model}

Building on the cover of sieves and its matrix interpretation, we define the
\emph{Sieve Neural Network} ($\mathsf{SNN}$) in two variants of increasing
flexibility. A detailed comparison with MPNNs is provided in
Appendix~\ref{explanation_details}.

\paragraph{Variant $\mathsf{SNN}(\alpha,(l,k))$.}
In the $\alpha$ variant, for each ordered pair of nodes $(v_i,v_j)$ we interpret
$\mathsf{CoSieve}(v_i,l)$ as a \emph{sender} and $\mathsf{Sieve}(v_j,k)$ as a
\emph{receiver}. The admissible transmissions from $v_i$ to $v_j$ are precisely
the paths allowed by the composed element
\(
  \mathsf{CoSieve}(v_i,l)\;\bullet\; \mathsf{Sieve}(v_j,k),
\)
(see Figure~\ref{sieve}). Under the map $\mathsf{Tr}$, the number of such
paths equals the $(i,j)$ entry of
\(
  \mathsf{CoImage}(v_i,l)\;\circ\; \mathsf{Image}(v_j,k).
\)
To obtain a scale-aware score, we normalize by the \emph{sending capacity} of
$v_i$ and the \emph{receiving capacity} of $v_j$:
let
{\small\[
  r_i \;=\; \sum_{q} \big(\mathsf{CoImage}(v_i,l)\big)_{iq},
  \qquad
  c_j \;=\; \sum_{p} \big(\mathsf{Image}(v_j,k)\big)_{pj}.
\]}
Then the output matrix of $\mathsf{SNN}(\alpha,(l,k))$ on $G$ is \(S^{(\alpha)}\) with $S^{(\alpha)}_{ij}
  \;=\;
  \frac{\big(\mathsf{CoImage}(v_i,l)\;\circ\;\mathsf{Image}(v_j,k)\big)_{ij}}
       {\,r_i\,c_j \,}$.
Intuitively, $S^{(\alpha)}_{ij}$ is the fraction of realized paths
relative to a node-pair–specific capacity. If the normalization is omitted, we
denote the model by $\mathsf{SNN}_o(\alpha,(l,k))$.

\paragraph{Variant $\mathsf{SNN}(\beta,(l_1,\dots,l_t))$.}
The $\beta$ variant aggregates information globally by summing over node-centered
images and co-images, alternating sender/receiver roles across depths. The model output is
\(
  S^{(\beta)} = Su_1 \circ Su_2 \circ \cdots \circ Su_t,
\)
where $Su_i = \sum_{v \in V} \mathsf{CoImage}(v,l_i)$ for odd $i$ and $Su_i = \sum_{v \in V} \mathsf{Image}(v,l_i)$ for even $i$, with $1 \leq i \leq t$.

\paragraph{How $\mathsf{SNN}$ is used.}
$\mathsf{SNN}$ is applied \emph{once} as a preprocessing step to transform each
graph: we replace its adjacency matrix with the corresponding $\mathsf{SNN}$
output (either $S^{(\alpha)}$ or $S^{(\beta)}$). The transformed dataset can then
be fed to any message-passing GNN, substituting the traditional
neighborhood cover with the sieve cover. The \emph{time complexity} of this
transformation is analyzed in Appendix~\ref{complex}.

\paragraph{Invariance.}
The model respects graph isomorphisms (node relabelings), making it suitable for
graph isomorphism and classification tasks.

\begin{theorem}\label{snn invariant}
$\mathsf{SNN}$ is invariant under graph isomorphism.
\end{theorem}

\subsection{Expressivity and WL}
The main theoretical contribution of this paper is the duality established between graphs and monoids. In Theorems~\ref{iso_graph-induce} and~\ref{induce_iso_graph} we show that the monoidal structures $\mathsf{Mod}(G)$ and $\mathsf{Mom}(G)$, together with the homomorphism $\mathrm{Tr}$, characterize graphs up to isomorphism.
Since GkGNN is an algebraic design framework rather than a bounded-dimensional color-refinement procedure, it is not formulated within the $k$-WL hierarchy. Instead, it introduces a different axis of expressivity grounded in monoidal compositions of directed subgraphs and their matrix realizations. For the sieve construction, we further prove that the submonoid generated by the cover of sieves also characterizes graphs up to isomorphism (Appendix~\ref{submonoid_of_sieves}). 

While the concrete model $\mathsf{SNN}$ realizes a restricted family of elements from this submonoid, we can establish a principled bridge between these algebraic path compositions and bounded-dimensional refinement by examining the initialization phase of the $k$-WL test. 

In the standard $k$-WL test, the initial hashing of node $k$-tuples plays an essential role in determining the algorithm's expressive bounds. This classical hashing mechanism is fundamentally based on the \textbf{cover of neighborhoods}, as it evaluates immediate adjacencies to capture the atomic isomorphism type of a tuple. Specifically, it requires identifying whether an edge exists between $v_i$ and $v_j$ for any $i,j$. Within our framework, this is algebraically equivalent to evaluating if there exists any valid path between $v_i$ and $v_j$ in the composed element $\mathsf{CoSieve}(v_i,1)\bullet\mathsf{Sieve}(v_j,0)$. Here, we adapt the $k$-WL algorithm to leverage the \textbf{cover of sieves} rather than standard neighborhoods. Inspired by this direct equivalence, we define a natural generalization:

\begin{definition}
    \textbf{Sieve-adapted $(r,s,k)$-WL algorithm.} The Sieve-adapted $(r,s,k)$-WL algorithm operates identically to the standard $k$-WL algorithm, with the exception that for each tuple $\mathbf{v}=(v_1,\cdots, v_k)$, its initial hash function is enriched to capture all paths between $v_i$ and $v_j$ residing within the algebraic element $\mathsf{CoSieve}(v_i,r)\bullet\mathsf{Sieve}(v_j,s)$, in addition to the tuple $\mathbf{v}$ itself.
\end{definition}

Under this definition, it is obvious that the $(1,0,k)$-WL algorithm, which restricts its view to the cover of neighborhoods, is strictly equivalent to the classical $k$-WL algorithm. Furthermore, based on the definition of the monoidal operation $\bullet$ and the construction of the elements $\mathsf{Sieve}(v_i,r)$ and $\mathsf{CoSieve}(v_j,s)$, one can easily verify a property of monotonicity: for any $r\le r'$ and $s\le s'$, the element $\mathsf{CoSieve}(v_i,r')\bullet\mathsf{Sieve}(v_j,s')$ inherently contains all paths present in $\mathsf{CoSieve}(v_i,r)\bullet\mathsf{Sieve}(v_j,s)$. This guarantees that increasing the sieve radii monotonically increases the topological distinguishing power of the initial tuple hashing.

Without simulating higher-order $k$-tuples, $\mathsf{SNN}$ leverages the expanded topological pathways of $\mathsf{CoSieve}(v_i,r)\bullet\mathsf{Sieve}(v_j,s)$ to natively capture structural information. Empirically, this leverage is evaluated by $\mathsf{SNN}$ achieving a 0\% failure rate on SRG families and the BREC benchmark (Table \ref{tab:sr_brec}), cleanly distinguishing non-isomorphic structures that bounded $k$-WL models cannot.

\begin{figure}[t]
    \centering
    \includegraphics[scale=0.55]{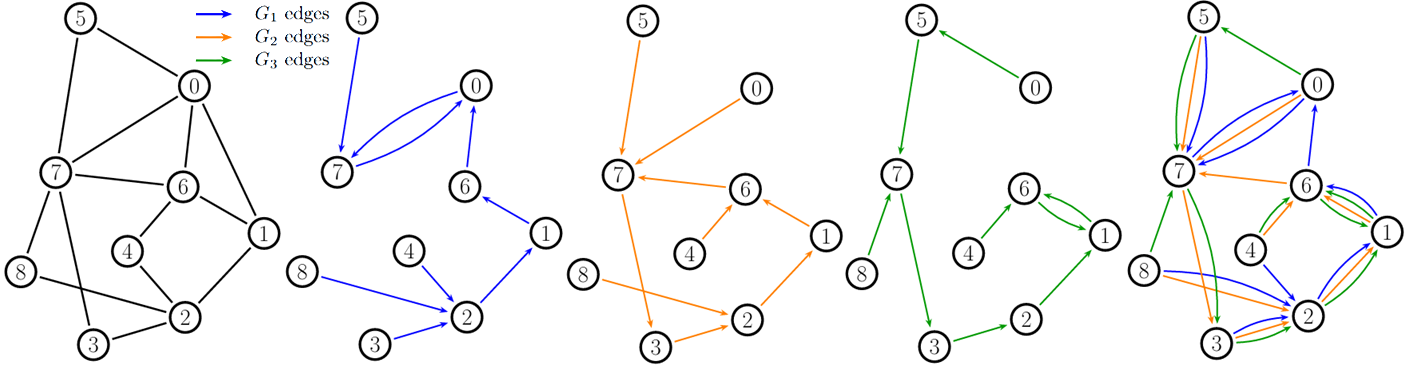}
    \caption{\textbf{Random walk--based monoidal construction.}
For a graph $G$ (left), we construct three directed multigraphs
$G_1$ (blue), $G_2$ (orange), and $G_3$ (green) by applying Steps~1--2 of
Example~\ref{random walk}. Their multigraph sum
$M = G_1 \oplus G_2 \oplus G_3$ is shown on the right.
Let $\hat G_1,\hat G_2,\hat G_3$ be as in Example~\ref{random walk}
(i.e., $\hat G_i=(G_i,E(G_i))$), and set
$(M,S)=\hat G_1 \bullet \hat G_2 \bullet \hat G_3 \in \mathsf{SMult}(G)$.
Then $S$ consists of admissible directed paths of length at most $3$ obtained by composing
\emph{at most one edge from each} $G_i$ in the order $G_1 \prec G_2 \prec G_3$:
length-$2$ paths are composable pairs with strictly increasing colors (blue$\to$orange,
blue$\to$green, or orange$\to$green), and length-$3$ paths are composable triples
blue$\to$orange$\to$green.
In particular, repeated colors (e.g., blue$\to$blue) and any backward color order
(e.g., orange$\to$blue or green$\to$orange) are not allowed.
This figure illustrates how the monoidal product $\bullet$ enforces ordered path composition.}
    \label{}
\end{figure}

\subsection{Further examples}

To illustrate the flexibility of the GkGNN framework in incorporating diverse structural principles—deterministic or stochastic—into unified message-passing strategies, we present two examples that are entirely distinct from the sieve-based construction of $\mathsf{SNN}$.

\begin{example}\label{random walk}
\textbf{Random Walk--Based Monoidal Elements.}

Let $G=(V,E)$ be a graph. We construct stochastic elements of
$\mathsf{SMult}(G)$ via a simple random-walk mechanism.

\emph{Step 1.}
For each node $v\in V$, independently and uniformly select a neighbor
$w\in N(v)$ and create a directed edge $(v\to w)$.

\emph{Step 2.}
Let $G_i$ be the directed multigraph consisting of all such edges.
We equip $G_i$ with the set of its directed edges, viewed as length-one
paths, and define
\[
\hat G_i := \bigl(G_i, E(G_i)\bigr)\in\mathsf{SMult}(G).
\]

Repeating this procedure for $t$ independent iterations yields
elements $\hat G_1,\ldots,\hat G_t\in\mathsf{SMult}(G)$.
Their monoidal composition
\[
\hat G_1 \bullet \hat G_2 \bullet \cdots \bullet \hat G_t
\]
encodes multi-step transition structure induced by successive random
walks. While each $G_i$ consists only of one-step transitions, longer
paths arise through the composition operation $\bullet$.

To implement this construction, let $A_1,\ldots,A_t$ be the adjacency
matrices of the directed multigraphs $G_1,\ldots,G_t$.
Applying the GkGNN matrix operation yields
\[
A_1 \circ A_2 \circ \cdots \circ A_t \in \mathsf{Mom}(G).
\]

The resulting matrix aggregates stochastic walk information and defines
a message-passing operator driven by randomized structural exploration.
\end{example}

\begin{example}\label{flow-based}
\textbf{Flow-Based Construction via Partitions.}

Let $G=(V,E)$ be a graph and suppose that $\{V_i\}_{i=1}^{k}$ is a partition of $V$, i.e.,
\[
\bigcup_{i=1}^{k} V_i = V,
\qquad
V_i \cap V_j = \varnothing \quad \text{for } i \neq j.
\]

For each block $V_i$, define
\[
\mathsf{Mask}(V_i) := 
\bigoplus_{v_j \in V_i} \mathsf{Tr}(N(v_j))^\top.
\]

Operationally, $\mathsf{Mask}(V_i)^\top$ is obtained from the adjacency matrix by zeroing out all rows corresponding to nodes not in $V_i$.
We then define
\[
F = \mathsf{Mask}(V_1) \circ \mathsf{Mask}(V_2) \circ \cdots \circ \mathsf{Mask}(V_k)
\in \mathsf{Mat}_{|V|}(\mathbb{R}).
\]

The operator $F$ encodes ordered information flow across partition blocks, from $V_1$ through $V_k$, and thus defines a structured propagation mechanism determined by the chosen partition.
In particular, contributions from nodes in $V_i$ can influence nodes in $V_j$ only through compositions consistent with the block order, thereby enforcing a coarse-grained flow of information across the graph.

Since the partition of $V$ may be defined arbitrarily (e.g., based on node degree, community structure, or a probability distribution over nodes), this construction illustrates how external structural criteria can be algebraically incorporated into message passing within the GkGNN framework.
In this sense, partitions of the node set naturally induce families of propagation operators, demonstrating that GkGNN can integrate clustering or community-level structure directly into the design of message-passing mechanisms.
\end{example}

\section{Experiments}\label{section 5}
To show how a shift in perspective improves graph understanding, we conduct a comprehensive evaluation of $\mathsf{SNN}$ on two tasks: (i) graph isomorphism and (ii) a topology-encoding probe.
\subsection{Graph isomorphism.}
We evaluate $\mathsf{SNN}$ on standard isomorphism benchmarks to test its ability
to distinguish non-isomorphic graphs and to compare against WL-style limits.
Throughout, we use the $\beta$ variants in strong (saturated) settings, and we do
\emph{not} train any parameters: each graph $G$ is mapped once to an
$\mathsf{SNN}$ output matrix $S(G)$, from which we compute simple permutation-invariant
summaries as embeddings. By invariance, isomorphic graphs yield identical embeddings;
non-identical embeddings imply non-isomorphism.\footnote{In practice we compare
embeddings with a small numerical tolerance. The depth ``$-1$'' denotes the
saturated sieve (Section~\ref{section_snn}).}

\paragraph{SR (Strongly Regular graphs).}
We use \textbf{all} publicly available collections of strongly regular graphs from
\href{https://web.archive.org/web/20211018181525/http://users.cecs.anu.edu.au/~bdm/data/graphs.html}{Brendan McKay's Graph Data (archived)}.
SR graphs are challenging since $3$-WL cannot fully distinguish them~\citep{sc}. For each collection, we apply $\mathsf{SNN}(\beta,(-1,-1,-1))$ to every graph $G$ and compute a $6$-dimensional embedding from $S(G)$:
\(
\big(\det(S),\,\mathrm{Min}(S),\mathrm{Mean}(S),\ \mathrm{Var}(S),\ \mathrm{Mean}(\mathrm{diag}(S)),\ \mathrm{Var}(\mathrm{diag}(S))\big).
\)
By invariance and Theorem~\ref{iso_graph-induce}, $\det(S)$ is preserved under relabeling, so isomorphic graphs match.
Within each collection, $\mathsf{SNN}$ assigns distinct embeddings to \emph{all}
graphs, yielding a $0\%$
failure rate; see Table~\ref{tab:sr_brec}.

\paragraph{CSL (Circular Skip Links).}
CSL contains $150$ 4-regular graphs partitioned into $10$ isomorphism classes
and is widely used to probe GNN expressivity~\citep{csl1,csl2}. We run
$\mathsf{SNN}(\beta,(-1))$ on each graph and use $\mathrm{Sum}(S)$ (sum of all
entries of $S$) as a permutation-invariant scalar embedding. The resulting values
perfectly separate the $10$ classes: graphs within a class share the same value;
graphs from different classes do not.

\paragraph{BREC.}
BREC~\citep{brec} contains $400$ pairs of non-isomorphic graphs divided into four
categories (60 Basic, 140 Regular, 100 Extension, and 100 CFI), with cases that
remain indistinguishable even under the $4$-WL test. For each graph $G$, we
apply $\mathsf{SNN}(\beta,(-1,-1,-1,-1))$ and construct the same type of embedding
as used in the \textbf{SR} experiment. Across all BREC pairs, $\mathsf{SNN}$
consistently assigns distinct embeddings to the two graphs in each pair, yielding
a $0\%$ failure rate (Table~\ref{tab:sr_brec}).

\begin{table*}
\centering
\caption{Left: Failure rates of 3-WL and SNN across Strongly Regular graphs. 
Right: Number of distinguished pairs on \textbf{BREC}. 
Baseline values from \cite{brec}.}
\vspace{0.5em}
\begin{minipage}[t]{0.48\textwidth}
%\centering
\resizebox{1.32\textwidth}{!}{
\begin{tabular}{lcc|lcc}
\toprule
\textbf{Graph Category} & \textbf{3-WL (\%)} & \textbf{SNN (\%)} &
\textbf{Graph Category} & \textbf{3-WL (\%)} & \textbf{SNN (\%)} \\
\midrule
SRG(25,12,5,6)  & 100 & 0 & SRG(36,15,6,6)   & 100 & 0 \\
SRG(26,10,3,4)  & 100 & 0 & SRG(37,18,8,9)   & 100 & 0 \\
SRG(28,12,6,4)  & 100 & 0 & SRG(40,12,2,4)   & 100 & 0 \\
SRG(29,14,6,7)  & 100 & 0 & SRG(65,32,15,16) & 100 & 0 \\
SRG(35,16,6,8)  & 100 & 0 &                  &     &   \\
SRG(35,18,9,9)  & 100 & 0 &                  &     &   \\
SRG(36,14,4,6)  & 100 & 0 &                  &     &   \\
\bottomrule
\end{tabular}}
\end{minipage}%
\hfill
\begin{minipage}[t]{0.35\textwidth}
%\centering
\resizebox{.85\textwidth}{!}{
\begin{tabular}{lcccc}
\toprule
Model & Basic & Reg. & Ext. & CFI \\
 & (60) & (140) & (100) & (100) \\
\midrule
3-WL    & 60 & 50  & 100 & 60 \\
SSWL-P  & 60 & 50  & 100 & 38 \\
I$^2$-GNN & 60 & 100 & 100 & 21 \\
GSN     & 60 & 99  & 95  & 0  \\
PPGN    & 60 & 50  & 100 & 23 \\
\midrule
$\mathsf{SNN}$ & 60 & 140 & 100 & 100 \\
\bottomrule
\end{tabular}}
\end{minipage}
\label{tab:sr_brec}
\end{table*}

\subsection{Topology Encoding (probe).}
Our aim here is to evaluate \emph{only} the topology encoded by $\mathsf{SNN}$ as a preprocessing operator, independently of any learnable parameters or downstream training. To do so, we design a parameter-free \emph{probe} based on one-step Label Propagation (LP) \citep{lp1, lp2} with $\alpha=1$. This choice isolates the structural signal present in the propagation operator and avoids confounds from optimization, regularization, or model capacity.

We compare two families of propagation operators. The first consists of classical adjacency-based operators, including the adjacency matrix $Ad$ and polynomial variants such as $Ad^2$, $Ad+Ad^2$, and $Ad+Ad^2+Ad^3$. The second family consists of operators generated by $\mathsf{SNN}$. Concretely, we run LP on
\(
A\in\{ Ad,\ Ad^2,\ Ad+Ad^2,\ Ad+Ad^2+Ad^3,\ \mathsf{SNN}(\beta,(1,1)),\ \mathsf{SNN}(\beta,(1,1,1)),\ \mathsf{SNN}(\beta,(1,2)),\ \mathsf{SNN}(\beta,(2,2))\}.
\)

\medskip
\medskip
\noindent\textbf{LP update (one step, no learning).}
Given initial one-hot labels $Y^{(0)}$ on the training nodes (zeros elsewhere), we perform a single propagation step
\(
Y^{(1)} = \widehat{M} Y^{(0)} .
\)
Here $\widehat{M}$ is a normalized version of the chosen operator $A$. We report results for three standard normalizations
\(
\widehat{M} \in
\left\{
D^{-1/2} A D^{-1/2},\;
D^{-1}A,\;
AD^{-1}
\right\},
\)
denoted respectively as \texttt{DAD}, \texttt{DA}, and \texttt{AD}, where $D$ is the degree matrix induced by $A$. This ensures comparability across covers and conforms to standard LP practice.
We evaluate on \textbf{Cora}, \textbf{CiteSeer}, \textbf{PubMed}, \textbf{ogbn-arxiv}, \textbf{AmazonPhoto}, and \textbf{Actor}; the dataset specifications and the runtimes of models $\mathsf{SNN}(\beta,(1,1))$ and $\mathsf{SNN}(\beta,(1,1,1))$ are reported in Table~\ref{tab:snn_runtime}. Because LP has no learnable parameters, differences in accuracy primarily reflect the structural information encoded by the propagation operator $A$.

\paragraph{Discussion.}
Table~\ref{tab:topology_encoding} compares classical polynomial propagation operators with operators derived from sieve covers under the same one-step label propagation probe. Because the probe uses a single propagation step and no learnable parameters, differences in accuracy reflect only the structural information encoded by the propagation matrix.

Among classical operators, increasing the degree of the adjacency polynomial often improves performance on several citation-style datasets. For example, on Cora and CiteSeer, moving from the adjacency matrix to $Ad^2$, $Ad + Ad^2$, and $Ad + Ad^2 + Ad^3$ gradually increases accuracy, suggesting that incorporating longer walks allows the operator to capture broader connectivity patterns.

However, increasing polynomial depth does not always lead to further improvements. On \textit{ogbn-arxiv}, the best classical operator is $Ad+Ad^2$ (0.660), while adding the third power of $Ad$ slightly decreases performance. This suggests that simply increasing the propagation radius does not necessarily provide additional useful structural information on larger graphs.

The Actor dataset provides a particularly clear contrast between the two operator families. Classical polynomial operators remain close to the adjacency baseline (around $0.19$), whereas sieve-based operators reach $0.254$. Since the probe does not involve training, this improvement reflects differences in the topology encoded by the propagation operators themselves.
The results on Actor also indicate that the specific structure of the sieve can influence performance. For instance, $\mathsf{SNN}(\beta,(1,2))$ achieves higher accuracy than $\mathsf{SNN}(\beta,(1,1,1))$.

Overall, these results illustrate that polynomial adjacency operators can capture useful multi-hop connectivity patterns, while operators derived from sieve covers may encode additional structural information in certain graphs. In particular, the Actor dataset highlights a setting where this difference becomes visible even under a single propagation step.

\begin{table*}[t]
\centering
\caption{Test accuracy of one-step Label Propagation ($\alpha=1$) using different propagation operators. The upper part reports classical polynomial adjacency operators, while the lower part reports operators derived from SNN. Since the probe uses a single propagation step and no learnable parameters, differences in accuracy reflect the structural information encoded by the propagation operator. The best classical baseline is underlined and the best overall result is shown in bold.}
\label{tab:topology_encoding}
\small
\setlength{\tabcolsep}{4pt}

%=====================
% Classical operators
%=====================
\begin{tabular}{lccc|ccc|ccc|ccc}
\toprule
& \multicolumn{3}{c|}{$Ad$}
& \multicolumn{3}{c|}{$Ad^2$}
& \multicolumn{3}{c|}{$Ad + Ad^2$}
& \multicolumn{3}{c}{$Ad + Ad^2 + Ad^3$} \\
\cmidrule(lr){2-4}
\cmidrule(lr){5-7}
\cmidrule(lr){8-10}
\cmidrule(lr){11-13}

\textbf{Dataset}
& DAD & DA & AD
& DAD & DA & AD
& DAD & DA & AD
& DAD & DA & AD \\
\midrule

Cora
& 0.260 & 0.260 & 0.258
& 0.485 & 0.483 & 0.477
& 0.514 & 0.508 & 0.506
& \underline{0.607} & 0.600 & 0.600 \\

CiteSeer
& 0.137 & 0.137 & 0.137
& 0.229 & 0.225 & 0.227
& 0.257 & 0.254 & 0.254
& \underline{0.377} & 0.368 & 0.373 \\

PubMed
& 0.189 & 0.189 & 0.189
& 0.242 & 0.242 & 0.240
& 0.248 & 0.248 & 0.246
& 0.423 & \underline{0.428} & 0.422 \\

ogbn-arxiv
& 0.617 & 0.597 & 0.613
& 0.634 & 0.613 & 0.595
& \underline{0.660} & 0.640 & 0.619
& 0.637 & 0.636 & 0.567 \\

AmazonPhoto
& 0.337 & 0.337 & 0.344
& 0.742 & 0.738 & 0.736
& 0.742 & 0.738 & 0.739
& \underline{\textbf{0.827}} & 0.792 & 0.761 \\

Actor
& 0.195 & 0.196 & 0.191
& 0.193 & 0.188 & 0.191
& 0.191 & 0.190 & 0.190
& 0.190 & \underline{0.195} & 0.191 \\

\bottomrule
\end{tabular}

\vspace{6pt}

%=====================
% SNN operators
%=====================
\begin{tabular}{lccc|ccc|ccc|ccc}
\toprule
& \multicolumn{3}{c|}{$\mathsf{SNN}(\beta,(1,1))$}
& \multicolumn{3}{c|}{$\mathsf{SNN}(\beta,(1,2))$}
& \multicolumn{3}{c|}{$\mathsf{SNN}(\beta,(1,1,1))$}
& \multicolumn{3}{c}{$\mathsf{SNN}(\beta,(2,2))$} \\
\cmidrule(lr){2-4}
\cmidrule(lr){5-7}
\cmidrule(lr){8-10}
\cmidrule(lr){11-13}

\textbf{Dataset}
& DAD & DA & AD
& DAD & DA & AD
& DAD & DA & AD
& DAD & DA & AD \\
\midrule

Cora
& 0.512 & 0.505 & 0.507
& 0.608 & 0.599 & 0.608
& 0.609 & 0.602 & 0.608
& \textbf{0.669} & 0.647 & 0.624 \\

CiteSeer
& 0.261 & 0.258 & 0.259
& 0.368 & 0.369 & 0.369
& 0.370 & 0.368 & 0.372
& 0.427 & 0.416 & \textbf{0.438} \\

PubMed
& 0.248 & 0.248 & 0.246
& 0.423 & 0.429 & 0.425
& 0.423 & 0.429 & 0.424
& 0.638 & \textbf{0.642} & 0.621 \\

ogbn-arxiv
& \textbf{0.663} & 0.645 & 0.625
& 0.649 & 0.645 & 0.621
& 0.647 & 0.642 & 0.581
& 0.641 & 0.642 & 0.617 \\

AmazonPhoto
& 0.743 & 0.739 & 0.741
& 0.817 & 0.789 & 0.770
& 0.826 & 0.796 & 0.766
& 0.808 & 0.774 & 0.702 \\

Actor
& 0.188 & 0.192 & 0.185
& 0.253 & \textbf{0.254} & 0.245
& 0.190 & 0.191 & 0.191
& 0.254 & 0.251 & 0.241 \\

\bottomrule
\end{tabular}

\end{table*}

\subsection{Graph classification}\label{sec-graph-classif}
We evaluate $\mathsf{SNN}$ on four datasets from the TUD benchmark suite: \textbf{NCI1}, \textbf{IMDB-B}, \textbf{IMDB-M}, and \textbf{PROTEINS}. Our model is compared against several GNNs, including GIN \citep{gin}, PPGNs \citep{ppgn}, GSN \citep{gsn}, TL-GNN \citep{tlgnn}, SIN \citep{sc}, and CIN \citep{cw}, as well as classical graph kernels such as the WL kernel \citep{wlkernel} and GNTK \citep{gntk}. For all datasets except \textbf{NCI1}, we employ $\mathsf{SNN}(\alpha, (1,1))$. Recognizing the need for a more expressive variant for \textbf{NCI1}, we utilize $\mathsf{SNN}(\alpha, (1,2))$.

Equation~\ref{description of image} defines the propagation operator through the matrices $\mathsf{Tr}(D_i(v))$. Multiplying $\mathsf{Tr}(D_i(v))$ by a constant $\gamma \in (0,1]$ enhances the model in a manner that is sensitive to the length of paths. In our experiments, we set $\gamma = 0.5$ in all cases. For other hyperparameters, see Table~\ref{tab:hyperparams}. We treat the output of $\mathsf{SNN}$ as a weighted graph that is subsequently processed by a standard GNN layer. We utilize the GNN operator \textbf{GraphConv} provided by PyTorch Geometric \citep{pygeo}, based on the model introduced in \cite{graphconv}. Tenfold cross-validation is performed. The results, presented in Table~\ref{results}, show that $\mathsf{SNN}$ achieves good performance across this diverse set of datasets.
We note that several methods exhibit overlapping standard deviations in Table~\ref{results}. 
Since fold-level results for prior work are not available, formal statistical tests 
(e.g., t-tests) cannot be conducted. However, the observed differences between methods 
are often comparable in magnitude to the reported standard deviations, suggesting that 
these improvements should be interpreted with caution. Accordingly, Table~\ref{results} 
is intended to highlight overall performance trends rather than statistically verified superiority.

\begin{table*}\smaller
\centering
\caption{Accuracy on TUD datasets. The best result for each dataset is shown in bold and the second best is underlined. $^\star$Graph Kernel Methods}
\begin{tabular}{lcccc}
\toprule
Dataset & NCI1 & IMDB-B & IMDB-M & PROTEINS \\
\midrule
WL kernel$^\star$ \citep{wlkernel} & \textbf{86.0±1.8} & 73.8±3.9 & 50.9±3.8 & 75.0±3.1 \\

GNTK$^\star$ \citep{gntk} & \underline{84.2±1.5} & 76.9±3.6 & 52.8±4.6 & 75.6±4.2 \\

GIN \citep{gin} & 82.7±1.7 & 75.1±5.1 & 52.3±2.8 & 76.2±2.8 \\

PPGNs \citep{ppgn} & 83.2±1.1 & 73.0±5.8 & 50.5±3.6 & 77.2±4.7 \\

GSN \citep{gsn} & 83.5±2.0 & 77.8±3.3 & 54.3±3.3 & 76.6±5.0 \\

TL-GNN \citep{tlgnn} & 83.0±2.1 & \underline{79.7±1.9} & \textbf{55.1±3.2} & \textbf{79.9±4.4} \\

SIN \citep{sc} & 82.8±2.2 & 75.6±3.2 & 52.5±3.0 & 76.5±3.4 \\

CIN \citep{cw} & 83.6±1.4 & 75.6±3.7 & 52.7±3.1 & \underline{77.0±4.3} \\

\midrule
SNN & 83.6±1.2 & \textbf{80.5±3.0} & \underline{54.53±2.23} & \underline{78.70±4.29} \\
\bottomrule
\end{tabular}
\label{results}
\end{table*}

\paragraph{Remarks on heterophilic graphs.}
Many real-world graphs exhibit heterophily, where neighboring nodes often
belong to different classes. In such settings, propagation based purely on
immediate neighborhoods may be less informative. The GkGNN framework allows
propagation operators to be constructed from structured compositions of
paths rather than relying solely on adjacency. In particular, $\mathsf{SNN}$ aggregates information along direction-respecting paths induced by
the chosen cover.
For illustration, we report results on the Actor dataset in Table \ref{tab:topology_encoding}, which
is commonly regarded as heterophilic. In this experiment we apply one-step
label propagation ($\alpha=1$) on operators derived from $\mathsf{SNN}$ without any
training. Using the operator $\mathrm{SNN}(\beta,(1,2))$ as a preprocessing
transformation yields $25.4\%$ accuracy. For reference, the performance of a
trained GCN baseline on this dataset is reported as $26.86\%$ in
\cite{geom-gcn}.
More generally, the GkGNN framework allows propagation operators to
incorporate external structural criteria. For example,
Example~\ref{flow-based} illustrates how a partition of the node set can
induce a structured flow operator that constrains how information
propagates across groups of nodes.

\paragraph{Limitations.}
GkGNN is a general design framework rather than a single model, with an intentionally
broad design space. Consequently, practical limitations depend on how the framework
is instantiated through specific cover constructions. In this work, we therefore
discuss limitations primarily at the level of $\mathsf{SNN}$,
which serves as an instantiation.
In our experiments, we focus on $\mathsf{SNN}$.
We observe that highly saturated variants of $\mathsf{SNN}$, while effective at exposing
structural differences between non-isomorphic graphs, may be overly discriminative
for certain graph classification tasks. In particular, saturation can amplify
fine-grained structural distinctions to the extent that similar graphs within the
same class become less aligned in the induced representations, which may reduce
classification robustness in downstream tasks.
Another practical limitation is that saturated sieve constructions often produce
dense matrices. While this is acceptable for offline analysis and expressivity
probes, practical deployment may require sparsification or masking strategies to
control computational and memory costs. As discussed in Subsection \ref{sec-graph-classif}, the output of
$\mathsf{SNN}$ can be enhanced to become sensitive to the length of paths. In such
settings, graph sparsification techniques could potentially be applied to control
the density of the resulting operators.
For example, spectral sparsification methods such as \cite{spars-siam}
aim to approximate dense graphs with sparser ones while preserving
important structural properties. More broadly, a range of graph reduction
techniques—including sparsification, coarsening, and condensation—have
been studied to control computational costs in graph learning pipelines
\citep{graph_reduction_survey}. 
In practice, sparsification may be applied at different stages of the
pipeline. For graphs that are already dense, it may be preferable to
first sparsify the graph before constructing the $\mathsf{SNN}$ operator,
in order to prevent excessive density in the resulting matrices. Recent
learning-based sparsification approaches also explore constructing
task-adaptive sparse graphs for scalable GNN computation
\citep{mog_sparsification}. 
In addition, stochastic edge-thinning approaches such as
\textit{DropEdge} \citep{spars_drop-edge}, which randomly remove edges
during training, might also be considered when $\mathsf{SNN}$ operators
are used within learning pipelines.
Exploring such strategies within the GkGNN
framework is left for future work.

\section{Conclusion}
We formalized \emph{covers} as a strict algebraic extension of neighborhoods and, in doing so, established a new foundation for message passing in which neighborhoods are recovered as a special case. We introduced the GkGNN framework to systematically construct covers and translate them into matrices, providing a principled mechanism for designing message-passing operators beyond adjacency-based aggregation. This platform simplifies the development of topology-aware propagation schemes. As a concrete instantiation, we proposed Sieve Neural Networks (SNN), which operationalize the framework and demonstrate strong performance on graph isomorphism and topology-encoding probes.
This work focuses on the foundational layer of GkGNN. Looking ahead, we will deepen the theoretical analysis of GkGNN’s expressive power and study how covers can be parameterized, learned, and adapted within end-to-end architectures. We also plan to broaden the scope of applications, including a more comprehensive theoretical and empirical comparison between SNN, GkGNN-based models, and the Weisfeiler–Lehman hierarchy.

\bibliography{main}
\bibliographystyle{tmlr}

\appendix

\section{Definitions and examples}\label{defapp}
\subsection{Definitions}
The definition of a monoid and monoidal homomorphism are as follows \citep{hun}:
\begin{definition}\label{def_monoid}
    A monoid is a non-empty set $\mathsf{M}$ together with a binary operation $\cdot$ on $\mathsf{M}$ which
    \begin{itemize}
        \item[1)] is associative: $a\cdot(b\cdot c)=(a\cdot b)\cdot c$ for all $a,b,c\in \mathsf{M}$ and

   \item[2)] contains identity element $e\in\mathsf{M}$ such that $a\cdot e=e\cdot a=a$
    \end{itemize}
   If, for all $a, b \in \mathsf{M}$, the operation satisfies $a \cdot b = b \cdot a$, then we say that $\mathsf{M}$ is a commutative monoid.
\end{definition}

\begin{definition}
A \emph{monoid homomorphism} between monoids $(M,\bullet)$ and $(N,\circ)$ 
with identity elements $e_M$ and $e_N$, respectively, 
is a function $f : M \to N$ such that
\[
f(x \bullet y) \;=\; f(x) \circ f(y) \quad \text{for all } x,y \in M,
\qquad f(e_M) = e_N.
\]
\end{definition}

\subsection{Examples}

\begin{example}\label{exco}
Considering a Change-of-Order mapping $f:\mathsf{Mat}_{3}(\mathbb{R})\rightarrow\mathsf{Mat}_{3}(\mathbb{R})$, obtained by reordering the standard basis $\lbrace e_1, e_2, e_3\rbrace$ to the basis  $\{ e_3, e_2, e_1\}$. For a given matrix $A$, we get the matrix $f(A)$ as follows:
\[A\longmapsto f(A)\]
  \[\bordermatrix{%
   & e_1 & e_2 & e_3 \cr
 e_1 & a_{11} & a_{12}& a_{13} \cr
 e_2 & a_{21} & a_{22}& a_{23} \cr
 e_3 & a_{31} & a_{32}& a_{33}
 } \xymatrix{\ar@{|->}[r]^{f:e_1\leftrightarrow e_3}&} 
 \bordermatrix{%
   & e_3 & e_2 & e_1 \cr
 e_3 & a_{33} & a_{32}& a_{31} \cr
 e_2 & a_{23} & a_{22}& a_{21} \cr
 e_1 & a_{13} & a_{12}& a_{11}
 }\]
\end{example}
\section{Explanation for constructing a model in GkGNN framework}\label{expofggnn}
The process of designing a GNN model within this framework is outlined as follows:
\begin{itemize}
    \item[1)] For a given graph $G$, the process involves selecting a collection $\mathcal{C}_G$ of elements from $\mathsf{Mod}(G)$ to serve as a cover for $G$. These elements can be generated using $\mathsf{DirSub}(G)$ and the binary operation $\bullet$. Notably, Theorem \ref{basis} ensures the ability to create any suitable and desired elements by leveraging directed edges and the operator $\bullet$.
    \item[2)] Next, the chosen cover is transformed into a collection of matrices within $\mathsf{Mom}(G)$, utilizing $\mathsf{Tr}$. During this transformation, the operation $\circ$ and other elements of $\mathsf{Mom}(G)$ can be employed to convert the original collection into a new one. The resulting output at this stage is denoted by $\mathcal{A}_G$.
    \item[3)] By utilizing $\iota$, the collection obtained in the second stage transitions into a larger and more equipped space, a suitable environment for enrichment. This stage leverages all the operations outlined in Proposition \ref{changeindices} to complete the model's design. Following the processing of $\mathcal{A}_G$ in this stage, we obtain a new collection of matrices denoted by $\mathcal{M}_G$, representing the model's output.
\end{itemize}
Hence, a model is a mapping that associates a collection of matrices $\mathcal{M}_G$ with a given graph $G$. $\mathcal{M}_G$ plays a role akin to the adjacency matrix and provides an interpretation of the chosen cover for use in various forms of message passing. While the second and third stages can be merged, we prefer to emphasize the significance of $\mathsf{Tr}$ in this process.

This construction of a model is appropriate for tasks such as node classification. For graph classification, we need an invariant construction. Based on Theorem \ref{iso_graph-induce}, a graph isomorphism $f: G\rightarrow H$ transform the triple $(\mathcal C_G,\mathcal A_G,\mathcal M_G)$ to a triple $(\mathcal C'_H,\mathcal A'_H,\mathcal M'_H)$ for graph $H$ and this may be different from $(\mathcal C_H,\mathcal A_H,\mathcal M_H)$. So a model constructed in the GkGNN framework is invariant if for every graph isomorphism $f: G\rightarrow H$, the maps $\mathsf{Mod}(f)$, $\mathsf{Mom}(f)$ and $\mathsf{CO}(f)$ induce one-to-one correspondences between $\mathcal C_G$ and $\mathcal C_H$, $\mathcal A_G$ and $\mathcal A_H$, and $\mathcal M_G$ and $\mathcal M_H$, respectively. The model $\mathsf{SNN}$ is an example of an invariant model.

\section{GkGNN framework vs.\ higher-order GNNs: a comparison}\label{comparison_ggnn_higher_orders}
We contrast GkGNN with higher-order GNNs such as MPSN~\cite{sc}, CWN~\cite{cw}, GSN~\cite{gsn}, and TLGNN~\cite{tlgnn}.

\paragraph{Framework, not a single model.}
GkGNN is a \emph{design framework}: it gives precise, graph-agnostic definitions of \emph{covers} (generalizing neighborhoods) and a principled way to turn them into matrices. Whereas higher-order GNNs typically hard-code one specific alternative to neighborhood aggregation, GkGNN provides an \emph{infinite design space} of covers, of which the standard neighborhood cover is a special case, enabling diverse message-passing strategies tailored to a task.

\paragraph{Topology-aware by construction.}
By Theorems~\ref{iso_graph-induce} and~\ref{induce_iso_graph}, GkGNN yields an algebraic description of a graph that is unique up to isomorphism. Each monoidal element of $\mathsf{Mod}(G)$ encodes concrete topological relationships; choosing a cover selects which aspects of topology to expose to downstream GNNs. Moreover, the algebra (composition, translation to matrices) lets one combine ideas from other paradigms within a single coherent toolkit.

\paragraph{Example: recovering $k$-hop message passing.}
GkGNN can reproduce common higher-order behaviors. Starting from the neighborhood cover $\{S_v : v\in G\}$, define for a node $v_k$ the set
\[
2\text{-}\mathsf{hop}(v_k)
= \big\{\, S_{v_{k_i}}\ \bullet\ e_i \;:\; v_{k_i}\in N(v_k),\ e_i: v_{k_i}\to v_k \,\big\}.
\]
Let $2\text{-}\mathsf{hop}(G)=\bigcup_{v_k} 2\text{-}\mathsf{hop}(v_k)$. Applying $\mathsf{Tr}$ maps this cover to a collection of matrices, which can be aggregated (e.g., by summation) to obtain a 2-hop propagation operator, mirroring the effect of $k$-hop message passing in~\cite{khop}.

\section{Further details on \textbf{SNN}}\label{explanation_details}

\subsection{Model explanations}
The $\mathsf{SNN}$ construction provides two ways to collapse the matrix collection induced by the sieve cover into a single operator: the \emph{$\alpha$}- and \emph{$\beta$}-variants.

\paragraph{$\alpha$-variant.}
Using $\mathsf{CoImage}(v,l)=\mathsf{Image}(v,l)^\top$, we obtain
\[
\mathsf{CoImage}(v_i,l)\circ \mathsf{Image}(v_j,k)
= \big(\,\mathsf{CoImage}(v_j,k)\circ \mathsf{Image}(v_i,l)\,\big)^\top.
\]
Hence the output of $\mathsf{SNN}(\alpha,(l,k))$ is the transpose of the output of $\mathsf{SNN}(\alpha,(k,l))$, and $\mathsf{SNN}(\alpha,(l,l))$ is symmetric. For $l\neq k$, symmetry need not hold (cf.\ Example~\ref{exmpnn}), so $\mathsf{SNN}(\alpha,(l,k))$ and $\mathsf{SNN}(\alpha,(k,l))$ may differ. Moreover, increasing the radii only adds admissible paths: if $l\le l'$ and $k\le k'$, then $\mathsf{SNN}(\alpha,(l',k'))$ captures (entrywise) at least as many paths as $\mathsf{SNN}(\alpha,(l,k))$.

\paragraph{$\beta$-variant.}
The families $\{\mathsf{Sieve}(v,l_i)\}_v$ (or $\{\mathsf{CoSieve}(v,l_i)\}_v$) form subcovers of the cover of sieves. Their matrix summaries
\[
Su_i \;=\; \sum_{v\in V}\mathsf{Image}(v,l_i)\quad\text{or}\quad
Su_i \;=\; \sum_{v\in V}\mathsf{CoImage}(v,l_i)
\]
aggregate all allowed paths contributed by the chosen subcover. Composing these summaries with the monoid operation $\,\circ\,$ produces
\[
Su_1 \circ \cdots \circ Su_t,
\]
which realizes a specific combination of subcovers: paths admitted by earlier subcovers are composed with those of later ones. Because $\,\circ\,$ is, in general, noncommutative, the order of $Su_i$ reflects the intended sequencing of interactions encoded by the cover.

\subsection{Comparing with MPNN}\label{comparing_mpnn}
For a node $v$, its neighborhood can be described by the element $\mathsf{Sieve}(v,1)$. Consequently, $\mathsf{SNN}_o(\alpha,(0,1))$ and $\mathsf{SNN}_o(\alpha,(1,0))$ correspond to the adjacency matrix, signifying their utilization of neighborhoods for message passing. This is equivalent to MPNNs. Hence, $\mathsf{SNN}$ can be considered as a generalization of MPNNs. In the following example, two graphs are considered that MPNN can not distinguish, yet $\mathsf{SNN}$ can. This example illustrates how a shift in perspective, resulting from a change in cover, reveals the topological properties of graphs.
\begin{example}\label{exmpnn}
The graphs in Figure \ref{mpnn cannot dis} are not distinguishable by MPNN \citep{sato} because they are locally the same.
\begin{figure}[h!]
    \centering
    \includegraphics[scale=0.3]{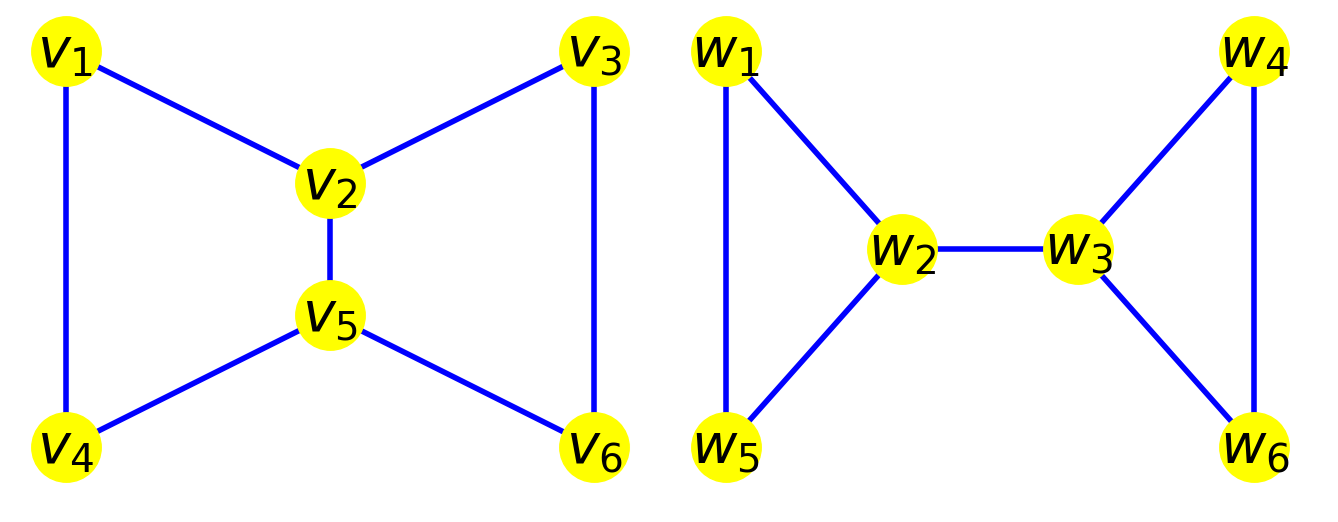}
    \caption{The graph $G$, the left one, and $H$, the right one, are not distinguishable by MPNN}
    \label{mpnn cannot dis}
\end{figure}
%\begin{figure}[h!]
 %   \centering
 %   \includegraphics[scale=0.2]{two graphs.png}
  % $$ \xymatrix{&&\\
   % v_1\ar@{-}[r]\ar@{-}[d] & v_2\ar@{-}[r]\ar@{-}[d]& v_3\ar@{-}[d]\\
   % v_4\ar@{-}[r]&v_5\ar@{-}[r] &v_6} \quad\quad\xymatrix{w_1\ar@{-}[dd]\ar@{-}[rd]& &&w_2\ar@{-}[dd]\ar@{-}[ld]\\
    %&w_3\ar@{-}[r]\ar@{-}[ld]&w_4\ar@{-}[rd]&\\
    %w_5&&&w_6}$$
    %\caption{The graph $G$, the left one, and $H$, the right one, are not distinguishable by MPNN}
    %\label{mpnn cannot dis}
%\end{figure}
Applying $\mathsf{SNN}_o(\alpha,(1,1))$, a level of version $\alpha$ of $\mathsf{SNN}$ that is slightly more potent than MPNN, we get the following symmetric matrices $X$ and $Y$ for $G$ and $H$ respectively as the outputs of the model for these graphs.
\[X=\begin{pmatrix}
    2& 2& 1& 2& 2& 0\\
    2& 3& 2& 2& 2& 2\\
    1& 2& 2& 0& 2& 2\\
    2& 2& 0& 2& 2& 1\\
    2& 2& 2& 2& 3& 2\\
    0& 2& 2& 1& 2& 2
\end{pmatrix} 
Y=\begin{pmatrix}
    2& 3& 1& 0& 3& 0\\
    3& 3& 2& 1& 3& 1\\
    1& 2& 3& 3& 1& 3\\
    0& 1& 3& 2& 0& 3\\
    3& 3& 1& 0& 2& 0\\
    0& 1& 3& 3& 0& 2
\end{pmatrix}\]
The entry $ij$ in these matrices corresponds to the count of paths between nodes $v_i$ and $v_j$ in $\mathsf{CoSieve}(v_i,1)\bullet\mathsf{Sieve}(v_j,1)$ and $w_i$ and $w_j$ in $\mathsf{CoSieve}(w_i,1)\bullet\mathsf{Sieve}(w_j,1)$. The disparity between these matrices highlights the differences between the graphs. This dissimilarity becomes more apparent when applying the set function $\mathsf{Var}$, while $\mathsf{Sum}$ and $\mathsf{Mean}$ yield identical values. When $\mathsf{SNN}_o(\alpha,(1,2))$, a more complex level of $\mathsf{SNN}$, is applied, we obtain the following nonsymmetric matrices, denoted as $Z$ and $W$, for graphs $G$ and $H$. Applying all three set functions results in distinct outputs, further emphasizing the dissimilarity between the graphs.
\[Z=\begin{pmatrix}
    2& 4& 2& 4& 4& 3\\
    5& 3& 5& 4& 6& 4\\
    2& 4& 2& 3& 4& 4\\
    4& 4& 3& 2& 4& 2\\
    4& 6& 4& 5& 3& 5\\
    3& 4& 4& 2& 4& 2
\end{pmatrix} 
W=\begin{pmatrix}
    2& 3& 3& 1& 3& 1\\
    4& 3& 4& 2& 4& 2\\
    2& 4& 3& 4& 2& 4\\
    1& 3& 3& 2& 1& 3\\
    3& 3& 3& 1& 2& 1\\
    1& 3& 3& 3& 1& 2
\end{pmatrix}\]
\end{example}

%%%%%%%%%%%%%%%%%%%%%%%%%%%%%%%%%%%%%%%
\subsection{Complexity}\label{complex}
$\mathsf{SNN}$ is applied \emph{once} as a preprocessing step to convert each input graph (or a dataset of graphs) into its transformed counterpart; it is not used during training.

Let $G=(V,E)$ with $|V|=n$ and $|E|=m$. From Eq.~\eqref{description of image}, $\mathsf{Image}(v,k)$ is obtained by $k$ iterations of adjacency-based additions/multiplications. The cost depends on the configuration:
\begin{itemize}
\item \textbf{$\mathsf{SNN}(\beta,(1,\dots,1))$.} In this case $\mathsf{Image}(v,k)$ can be read off directly from the adjacency matrix (no matrix–matrix products), so each $Su_i$ equals the adjacency matrix. Hence computing $S^{(\beta)}$ is $\mathcal{O}(mn)$.
\item \textbf{$\mathsf{SNN}(\alpha,(l,k))$ or $\mathsf{SNN}(\beta,(l_1,\dots,l_t))$ with $k>1$ or some $l_{i_0}>1$.} These require matrix-based compositions; computing $\mathsf{Image}(v,k)$ for a single node costs $\mathcal{O}(mn)$, yielding $\mathcal{O}(mn^2)$ over all nodes.
\end{itemize}
Since $\mathsf{SNN}$ runs only once to produce the transformed graphs, its runtime is incurred offline and does not affect the training-time complexity of downstream GNNs.

\begin{table}[t]
\centering
\caption{Dataset statistics and runtime (in seconds) for constructing $\mathsf{SNN}$-transformed graphs.
Reported times correspond to building the propagation operator before applying Label Propagation.}
\vspace{0.5em}
\begin{tabular}{lcccccc}
\toprule
\textbf{Dataset} & \textbf{\#Nodes} & \textbf{\#Edges} & \textbf{\#Features} & \textbf{\#Classes} & 
\makecell{\textbf{Runtime} \\ $\mathsf{SNN}(\beta,(1,1))$} & 
\makecell{\textbf{Runtime} \\ $\mathsf{SNN}(\beta,(1,1,1))$} \\
\midrule
Cora        & 2,708   & 10,556    & 1,433 & 7  & 0.0525 & 0.0758 \\
CiteSeer    & 3,327   & 9,104     & 3,703 & 6  & 0.0189 & 0.0319 \\
PubMed      & 19,717  & 88,648    & 500   & 3  & 0.8055 & 1.3902 \\
ogbn-arxiv  & 169,343 & 1,166,243 & 128   & 40 & 6.75   & 9.38 \\
AmazonPhoto & 7,650   & 238,162   & 745   & 8  & 0.6560 & 3.4766 \\
Actor       & 7,600   & 30,019    & 932   & 5  & 0.0224 & 0.1543 \\
\bottomrule
\end{tabular}
\label{tab:snn_runtime}
\end{table}

\begin{table}[h]
\centering
\caption{Training hyperparameters used for each dataset. All experiments were conducted on Google Colab using only CPU and standard RAM.}
\begin{tabular}{lccccc}
\hline
\textbf{Dataset} & \textbf{Learning Rate} & \textbf{Batch Size} & \textbf{GNN Layers} & \textbf{Dropout} & \textbf{Hidden Channels} \\
\hline
NCI1      & $1\text{e}^{-3}$ & 32  & 4 & 0.5 & 32  \\
IMDB-B    & $1\text{e}^{-3}$ & 32  & 4 & 0.5 & 64  \\
IMDB-M    & $1\text{e}^{-3}$ & 32  & 4 & 0.5 & 128 \\
PROTEINS  & $1\text{e}^{-3}$ & 32  & 4 & 0   & 32  \\
\hline
\end{tabular}
\label{tab:hyperparams}
\end{table}

\newpage
\subsection{Algorithm}\label{algorithm}

\begin{algorithm}[h!]
\caption{Computing $\mathsf{Image}(v,k)$}
\begin{algorithmic}[1] % [1] to number each line
\State \textbf{Input:} node $v$, integer $k$
\State \textbf{Output:} $\mathsf{Image}(v,k)$
\State \textbf{Initialization:} $N_0(v)=\lbrace v\rbrace$, $\mathsf{Image}(v,0)=\textit{Zero matrix}$
\For{$i=1, \cdots, k$}
    \State $N_i(v)=\bigcup_{u\in N_{i-1}(v)} N(u)-\bigcup_{j=0}^{i-1}N_j(v)$
    \State $M_i(v)=\lbrace w\rightarrow u: wu\in E, w\in N_i(v), u\in N_{i-1}(v)\rbrace$
    %\State Assuming $M_i(v)$ as a directed graph
    \State $\mathsf{Tr}(D_i(v))=\sum_{e\in M_i(v)}\mathsf{Tr}(e)=$ The adjacency matrix of directed subgraph $M_i(v)$
    \State $\mathsf{Image}(v,i)=\mathsf{Tr}(D_i(v))\circ\mathsf{Image}(v,i-1)$
    
    %\If{some condition}
     %   \State Do something
    %\Else
     %   \State Do something else
    %\EndIf
\EndFor
\State \textbf{Return:} Final result
\end{algorithmic}
\end{algorithm}

\begin{algorithm}[!]
\caption{Computing $\mathsf{CoImage}(v,k)$}
\begin{algorithmic}[1] % [1] to number each line
\State \textbf{Input:} $\mathsf{Image}(v,k)$
\State \textbf{Output:} $\mathsf{CoImage}(v,k)$
%\State \textbf{Initialization:} Describe any initialization steps
%\For{each element in the dataset}
 %   \State Perform some operation
  %  \If{some condition}
   %     \State Do something
    %\Else
     %   \State Do something else
    %\EndIf
%\EndFor
\State $\mathsf{CoImage(v,k)}=$ Transpose of $\mathsf{Image(v,k)}$
\State \textbf{Return:} Final result
\end{algorithmic}
\end{algorithm}

\begin{algorithm}[!]
\caption{Computing $\mathsf{SNN}(\alpha,(l,k))$}
\begin{algorithmic}[1] % [1] to number each line
\State \textbf{Input:} $\mathsf{Image}(v,k)$ and $\mathsf{CoImage}(v,l)$ for all $v\in V$
\State \textbf{Output:} $\mathsf{SNN}(\alpha,(l,k))$
\State \textbf{Initialization:} $\mathsf{SNN}(\alpha,(l,k))=$ \textit{Zero matrix}
\For{$v_i\in V$}
    \For{$v_j\in V$}
        \State $A=\mathsf{CoImage}(v_i,l)\circ \mathsf{Image}(v_j,k)$
        \State $r=\mathsf{CoImage}(v_i,l)[i,:].sum()$, summation of $i-$th row
        \State $c=\mathsf{Image}(v_j,k)[:,j].sum()$, summation of $j-$th column
        \State $\mathsf{SNN}(\alpha,(l,k))_{i,j}=\frac{A_{i,j}}{r\cdot c}$ 
    \EndFor
\EndFor
\State \textbf{Return:} Final result
\end{algorithmic}
\end{algorithm}

\begin{algorithm}[!]
\caption{Computing $\mathsf{SNN}(\beta,(l,k))$}
\begin{algorithmic}[1] % [1] to number each line
\State \textbf{Input:} $\mathsf{Image}(v,k)$ and $\mathsf{CoImage}(v,l)$ for all $v\in V$
\State \textbf{Output:} $\mathsf{SNN}(\beta,(l,k))$
%\State \textbf{Initialization:} $\mathsf{SNN}(\alpha,(l,k))=$ \textit{Zero matrix}
\State $Su_1=\sum_{v\in V}\mathsf{CoImage}(v,l)$
\State $Su_2=\sum_{v\in V}\mathsf{Image}(v,k)$
\State $\mathsf{SNN}(\beta,(l,k))=Su_1\circ Su_2$
\State \textbf{Return:} Final result
\end{algorithmic}
\end{algorithm}
\section{Special submonoids}
\subsection{The submonoid generated by neighborhoods}\label{submonoid_of_neigh}
The cover of neighborhoods, as a subset of $\mathsf{Mod}(G)$, generates a submonoid. To formalize this, let $\mathsf{Neigh}(G) \subseteq \mathsf{Mod}(G)$ and $\mathsf{Adj}(G)\subseteq\mathsf{Mom}(G)$ denote the submonoids generated by the cover of neighborhoods and its matrix transformation, respectively. The following theorems illustrate how these submonoids provide an algebraic characterization of a graph.
It is straightforward to verify that for a graph isomorphism $f: G \to H$, the mappings $\mathsf{Mod}(f)$ and $\mathsf{Mom}(f)$ send elements of $\mathsf{Neigh}(G)$ and $\mathsf{Adj}(G)$ to elements of $\mathsf{Neigh}(H)$ and $\mathsf{Adj}(H)$, respectively. Thus, as a consequence of Theorem \ref{iso_graph-induce}, we have:
\begin{theorem}
    Every graph isomorphism $f: G\rightarrow H$ induces monoidal isomorphisms $\mathsf{Neigh}(f):\mathsf{Neigh}(G)\longrightarrow\mathsf{Neigh}(H)$ and $\mathsf{Adj}(f):\mathsf{Adj}(G)\rightarrow\mathsf{Adj}(H)$ such that the following diagram is commutative, where $\iota$ represents the inclusions.
    \begin{equation}\label{diag for mon neigh}
        \xymatrix{\mathsf{Neigh}(G)\ar[r]^{\mathsf{Tr}_G}\ar[d]_{\mathsf{Neigh}(f)}&\mathsf{Adj}(G)\ar[d]_{\mathsf{Adj}(f)}\ar[r]^{\iota}&\mathsf{Mat}_{|V_G|}(\mathbb{R})\ar[d]^{\mathsf{CO}(f)}\\
    \mathsf{Neigh}(H)\ar[r]_{\mathsf{Tr}_H}&\mathsf{Adj}(H)\ar[r]_{\iota}&\mathsf{Mat}_{|V_H|}(\mathbb{R})
    }
    \end{equation}
\end{theorem}
The converse of this theorem can be stated as follows:
\begin{theorem}\label{monoid adj}
    Suppose $G$ and $H$ are two graphs with $|V_G|=|V_H|=n$, and $f:\mathsf{Mat}_{n}(\mathbb{R})\rightarrow \mathsf{Mat}_{n}(\mathbb{R})$ is a Change-of-Order mapping. If the restriction of $f$ to $\mathsf{Adj}(G)$ yields an isomorphism to $\mathsf{Adj}(H)$, then $G$ and $H$ are isomorphic.
\end{theorem}
Consequently, the horizontal homomorphisms in Diagram \ref{diag for mon neigh} can serve as an algebraic description of the graph.
It demonstrates that the monoidal elements resulting from interactions between neighborhoods encapsulate richer information about the graph's topology. This suggests that the coverage of neighborhoods can be further enhanced by incorporating additional elements from $\mathsf{Neigh}(G)$.

\subsection{The submonoid generated by sieves}\label{submonoid_of_sieves}
The submonoid generated by the cover of Sieves fully determines the graph, as stated in the following two theorems. Let $\mathsf{Si}(G) \subseteq \mathsf{Mod}(G)$ and $\mathsf{Im}(G) \subseteq \mathsf{Mom}(G)$ denote the submonoids generated by the cover of sieves and its matrix transformation, respectively. As a direct consequence of Theorems \ref{coversieveinvariant} and \ref{iso_graph-induce}, we have:
\begin{theorem}
    Every graph isomorphism $f: G\rightarrow H$ induces monoidal isomorphisms $\mathsf{Si}(f):\mathsf{Si}(G)\longrightarrow\mathsf{Si}(H)$ and $\mathsf{Im}(f):\mathsf{Im}(G)\rightarrow\mathsf{Im}(H)$ such that the Diagram \ref{graph iso si im} is commutative, where $\iota$ represents the inclusions.
    \begin{equation}\label{graph iso si im}
        \xymatrix{\mathsf{Si}(G)\ar[r]^{\mathsf{Tr}_G}\ar[d]_{\mathsf{Si}(f)}&\mathsf{Im}(G)\ar[d]_{\mathsf{Im}(f)}\ar[r]^{\iota}&\mathsf{Mat}_{|V_G|}(\mathbb{R})\ar[d]^{\mathsf{CO}(f)}\\
    \mathsf{Si}(H)\ar[r]_{\mathsf{Tr}_H}&\mathsf{Im}(H)\ar[r]_{\iota}&\mathsf{Mat}_{|V_H|}(\mathbb{R})
    }
    \end{equation}
\end{theorem}
The converse of the above theorem can be stated as follows:
\begin{theorem}\label{Im iso graph iso}
    Suppose $G$ and $H$ are two graphs with $|V_G|=|V_H|=n$, and $f:\mathsf{Mat}_{n}(\mathbb{R})\rightarrow \mathsf{Mat}_{n}(\mathbb{R})$ is a Change-of-Order mapping. If the restriction of $f$ to $\mathsf{Im}(G)$ yields an isomorphism to $\mathsf{Im}(H)$, then $G$ and $H$ are isomorphic.
\end{theorem}
The horizontal morphisms in Diagram \ref{graph iso si im} provide a unique, up-to-isomorphism algebraic characterization of a graph. This can be served as the basis for the significant performance of the model $\mathsf{SNN}$ in the graph isomorphism task, as demonstrated in the experimental section.
\section{Proof of theorems}\label{section_proofs}
%\subsection{Proof of Proposition \ref{transitive}}
%\begin{proof}
%Let $v_i\le_D v_j$ and $v_j\le_D v_k$, so there are paths in $D$ from $v_i$ to $v_j$ and $v_j$ to $v_k$; hence the concatenation of these paths is a path in $D$ from $v_i$ to $v_k$ and then $v_i\le_D v_k$.
%\end{proof}
\subsection{Proof of Theorem \ref{phi}}
\begin{proof}
    Since $\mathsf{Rep}$ is surjective, it suffices to demonstrate that $\mathsf{Rep}$ is also injective, meaning that if $\mathsf{Rep}(D) = \mathsf{Rep}(D')$, then $D = D'$. According to the matrix representation definition, $\le_D = \le_{D'}$. For an edge $\xymatrix{v_i\ar[r]^{e}& v_j} $ in $D$, it implies $v_i\le_D v_j$, and consequently, $v_i\le_{D'} v_j$. Suppose $\xymatrix{v_i\ar[r]^{e}& v_j}$ is not a directed edge in $D'$. In that case, there must be a path in $D'$ traversing a node $v_k$ different from $v_i$ and $v_j$. This implies $v_i\leq_{D'} v_k$ and $v_k\leq_{D'} v_j$, and consequently, $v_i\leq_{D} v_k$ and $v_k\leq_{D} v_j$. Thus, there is a path in $D$ from $v_i$ to $v_j$ traversing $v_k$. However, this path is distinct from $\xymatrix{v_i\ar[r]^{e}& v_j}$, contradicting the definition of directed subgraphs. Therefore, $e$ is a directed edge in $D'$. Similarly, we can demonstrate that every edge in $D'$ also belongs to $D$ with the same direction. Thus, $D=D'$.
\end{proof}
\subsection{Proof of Theorem \ref{smult}}
\begin{proof}
    The empty graph is its identity element, and the associativity of $\bullet$ comes from the associativity of the composition of paths. The non-commutativity is explained in Example \ref{noncommutativity}.
\end{proof}
\subsection{Proof of Theorem \ref{basis}}
\begin{proof}
    Since directed subgraphs, together with the operation $\bullet$ generate the monoid  $\mathsf{Mod}(G)$, we just need to show that every directed subgraph can be formed by its directed edges using the operation $\bullet$. We will prove this by induction based on the number of edges. Let $D$ be a directed subgraph of $G$. There is nothing to prove if $D$ has just one directed edge. Suppose the number of edges in $D$ is $m$, and the statement is true for every directed subgraph with edges less than $m$;  Our task is to show that the statement holds for $D$ as well.

    Let $V_D$ be the set of nodes of $D$. Since $\le_D$ is transitive, $(V_D, \le_D)$ can be seen as a partially ordered set, implying the existence of maximal elements. A node is considered maximal if it is not the starting point of any path. Now, let  $v$ be a maximal node; we choose a directed edge $\xymatrix{w\ar[r]^{e}& v}$ in $D$ and remove it. The following three situations may occur:
    \begin{itemize}
        \item[1)] producing one directed subgraph $D'$: $D$ and $D'\bigoplus e$ have the same directed edges. Since $v$ is maximal, the paths of $D$ that pass $e$ have this directed edge as their terminal edge. Then
        \[\mathsf{Paths}(D)=\mathsf{Paths}(D')\star e\]
        This follows $D=D'\bullet e$. Based on the assumption, $D'$ can be created by its edges. Then, the statement is true for $D$.
        \item[2)] producing two components where one of them is an isolated node, and the other one is a directed subgraph $D'$: in this case, we first remove the isolated node and then, similar to the first case, we conclude that the statement is true for $D$. 
        \item[3)] producing two directed subgraphs $D'$ and $D''$ where $w\in D'$ and $v\in D''$: obviously $D$ and $D'\bigoplus e\bigoplus D''$ have the same directed edges. With an argument similar to the first part, the maximality of $v$ implies
        \[\mathsf{Paths}(D)=\mathsf{Paths}(D')\star\lbrace e\rbrace\star\mathsf{Paths}(D'')\]
         and then $D=D'\bullet e \bullet D''$. Now, by the assumption that $D'$ and $D''$ can be created by their edges, the statement is true for $D$.         
    \end{itemize}
\end{proof}
\subsection{Proof of Theorem \ref{matmon}}
\begin{proof}
Since the summation and multiplication of matrices are associative, the operation $\circ$ is associative. The zero matrix is the identity element of $\mathsf{Mat}_{n}(\mathbb{R})$ with respect to $\circ$.
\end{proof}

\subsection{Proof of Theorem \ref{monoidal surjection}}
To define a monoidal homomorphism between the monoids $(\mathsf{Mod}(G), \bullet)$ and $(\mathsf{Mom}(G), \circ)$ in such a way that it is an extension of the morphism $\mathsf{Rep}$, we first prove the following theorem which gives a good explanation of the monoidal operation $\circ$.
\begin{theorem}\label{representation}
    For $A_1, A_2, \cdots, A_k\in \mathsf{Mat}_{n}(\mathbb{R})$ with $k\in \mathbb{N}$ we have: 
    \[A_1\circ A_2 \circ \cdots \circ A_k =\sum_{i=1}^{k}A_i
    +\sum_{\sigma\in O(k,2)}A_{\sigma_1}A_{\sigma_2}+\cdots
    +\sum_{\sigma\in O(k,j)}A_{\sigma_1}\cdots A_{\sigma_j}+ \cdots
    +A_1 A_2\cdots A_k\]
    where $O(k, i)$ is the set of all strictly monotonically increasing sequences of $i$ numbers of $\lbrace1, \cdots, k\rbrace$
\end{theorem}
\begin{proof}
    We prove the statement by induction on $k$. For $k=2$, there is nothing to prove, which is clear from the definition. Let the statement be true for $k$; We will show it is true for $k+1$. The associativity of $\circ$ and the induction hypothesis imply: 
    \[A_1\circ A_2 \circ \cdots \circ A_k \circ A_{k+1}=
    (A_1\circ A_2 \circ \cdots \circ A_k) \circ A_{k+1}=\]
    \[(A_1\circ A_2 \circ \cdots \circ A_k)+A_{k+1}+(A_1\circ A_2 \circ \cdots \circ A_k)A_{k+1}=\]   
    \[\sum_{i=1}^{k}A_i + \cdots +\sum_{\sigma\in O(k,j)} A_{\sigma_1} \cdots A_{\sigma_j}+\cdots +A_1 A_2\cdots A_k+\]
    \[A_{k+1}+\]
    \[(\sum_{i=1}^{k}A_i + \cdots +\sum_{\sigma\in O(k,j)} A_{\sigma_1} \cdots A_{\sigma_j}+\cdots +A_1\cdots A_k)A_{k+1}\]
    \[=\sum_{i=1}^{k+1}A_i + (\sum_{i=1}^{k}A_iA_{k+1}+\sum_{\sigma\in O(k,2)}A_{\sigma_1}A_{\sigma_2})+ \cdots +\]
    \[(\sum_{\sigma\in O(k,j-1)} A_{\sigma_1} \cdots A_{\sigma_{j-1}}A_{k+1}+\sum_{\sigma\in O(k,j)} A_{\sigma_1} \cdots A_{\sigma_j})+\]
    \[\cdots + A_1\cdots A_kA_{k+1}=\]
    \[\sum_{i=1}^{k+1}A_i+\sum_{\sigma\in O(k+1,2)}A_{\sigma_1}A_{\sigma_2}+ \cdots+\sum_{\sigma\in O(k+1,j)}A_{\sigma_1}\cdots A_{\sigma_j}+\]
    \[\cdots +A_1 A_2\cdots A_kA_{k+1}\]
    
    Therefore the statement is true for $k+1$.
\end{proof}
Now, we prove Theorem \ref{monoidal surjection}.
\begin{proof}
Considering that $S=\mathsf{Paths}(D_{1})\star \cdots \star \mathsf{Paths}(D_{k})$, let $p=p_0 p_1 \cdots p_m\in S$ be a path from $v_i$ to $v_j$ that is obtained by composition of subpaths $p_0\in \mathsf{Paths}(D_{i_0}), \cdots, p_m\in \mathsf{Paths}(D_{i_m})$ and $1\le i_0\lneqq\cdots\lneqq i_m\le k$. The number of all such paths from $v_i$ to $v_j$ equals the $ij$ entry of the matrix $(A_{i_0}\cdots A_{i_m})$ that is a summand of $A$ as explained in Theorem \ref{representation}. So the number of all paths from $v_i$ to $v_j$ in $S$ equals the $ij$ entry of $A$. Therefore, the definition of $\mathsf{Tr}$ just depends on $S$ and is independent of the choice of $D_i$s. Then $\mathsf{Tr}$ is well-defined. Based on the definition, $\mathsf{Tr}$ is a monoidal homomorphism. 

Suppose $B\in\mathsf{Mom}(G)$, then there are some matrix representations  $B_1, \cdots, B_l$ in $\mathsf{MatRep}(G)$ such that $B=B_1\circ \cdots\circ B_l$. Since $\mathsf{Rep}$ is an isomorphism, there exist some directed subgraphs $C_1, \cdots, C_l$ such that $\mathsf{Rep}(C_i)=B_i$.  Now, by choosing $C=C_1\bullet\cdots\bullet C_l$, we obtain $\mathsf{Tr}(C)=B$, establishing that $\mathsf{Tr}$ is surjective.
 \end{proof}
 \subsection{Proof of Proposition \ref{changeindices}}
 \begin{proof}
As we explained, $f$ changes the order of rows and columns. Thus, it preserves element-wise and matrix multiplications. Since $f$ is also linear, we have
\begin{equation*}
    \begin{split}
        f(A\circ B) & =f(A+B+AB) \\
        & =f(A)+f(B)+f(AB)\\
        & = f(A)+f(B)+f(A)f(B)\\
        & =f(A)\circ f(B)
    \end{split}
\end{equation*}
and then $f$ preserves the operation $\circ$ and this property establishes $f$ as a monoidal isomorphism.
\end{proof}
\subsection{Proof of Theorem \ref{iso_graph-induce}}
\begin{proof}
    Since $f$ is a change in the order, it induces bijections $\mathsf{DirSub}(f)$ and $\mathsf{MatRep}(f)$ such that Diagram \ref{com_rep_graph_iso} commutes.
    
    \begin{equation}\label{com_rep_graph_iso}
        \xymatrix{\mathsf{DirSub}(G)\ar[r]^{\mathsf{Rep}}\ar[d]_{\mathsf{DirSub}(f)}&\mathsf{MatRep}(G)\ar[d]^{\mathsf{MatRep}(f)}\\
    \mathsf{DirSub}(H)\ar[r]_{\mathsf{Rep}}&\mathsf{MatRep}(H)}
    \end{equation}
Also, $f$ induces monoidal isomorphism $\mathsf{SMult}(f):\mathsf{SMult}(G)\rightarrow \mathsf{SMult}(H)$ that sends $(M,S)\mapsto(f(M),f(S))$. According to the commutativity of the squares in Diagram \ref{def_modf_algf}, isomorphisms $\mathsf{Mod}(f):\mathsf{Mod}(G)\rightarrow\mathsf{Mod}(H)$ and $\mathsf{Mom}(f):\mathsf{Mom}(G)\rightarrow\mathsf{Mom}(H)$ can be obtained by restricting $\mathsf{SMult}(f)$ to $\mathsf{Mod}(G)$ and $\mathsf{CO}(f)$ to $\mathsf{Mom}(G)$.
    \begin{equation} \label{def_modf_algf}
    \xymatrix{\mathsf{DirSub}(G)\ar[r]^{\mathsf{DirSub}(f)}\ar[d]&\mathsf{DirSub}(H)\ar[d]\\ \mathsf{SMult}(G)\ar[r]_{\mathsf{SMult}(f)}&\mathsf{SMult}(H)}\ \ \ \hfill
    \xymatrix{\mathsf{MatRep}(G)\ar[r]^{\mathsf{MatRep}(f)}\ar[d]&\mathsf{MatRep}(H)\ar[d]\\
    \mathsf{Mat}_{|V_G|}(\mathbb{R})\ar[r]_{\mathsf{CO}(f)}&\mathsf{Mat}_{|V_H|}(\mathbb{R})}  
    \end{equation}
    The commutativity of the right square in Diagram \ref{graph iso implies com} directly follows from the definition of $\mathsf{Mom}(f)$. As illustrated in Diagram \ref{com_rep_graph_iso}, the left square in Diagram  \ref{graph iso implies com} is shown to be commutative for the generators of monoids, establishing the commutativity of this square.
\end{proof}
\subsection{Proof of Theorem \ref{induce_iso_graph}}
\begin{proof}
We begin by demonstrating that $f$ establishes a one-to-one correspondence between the edges of $G$ and $H$. It is evident that a matrix with a single non-zero entry in either $\mathsf{Mom}(G)$ or $\mathsf{Mom}(H)$ corresponds to a matrix transformation of an element in $\mathsf{Mod}(G)$ or $\mathsf{Mod}(H)$, respectively, each representing a single directed edge.

For an edge $\xymatrix{v_i\ar@{-}[r]&v_j}$ in $G$, let $e$ be the directed edge $v_i\rightarrow v_j \in\mathsf{Mod}(G)$; then $A=\mathsf{Tr}_G(e)$ has one non-zero entry, and since $f$ is a linear isomorphism, $f(A)$ has one non-zero entry, and, based on the assumption, it belongs to $\mathsf{Mom}(H)$. So $f(A)$ is a matrix transformation of a directed edge $c:u_k\rightarrow u_l$ in $\mathsf{Mod}(H)$. Similarly, let $B\in\mathsf{Mom}(G)$ be the matrix transformation of $e':v_j\rightarrow v_i$ and  then $f(B)\in\mathsf{Mom}(H)$ is a matrix transformation of some directed edge $c':u_{l'}\rightarrow u_{k'}$ in $\mathsf{Mod}(H)$. Since $e$ can be followed by $e'$, $e\bullet e'$ has three paths. This implies $\mathsf{Tr}_G(e\bullet e')$ has three non-zero entries. On the other hand, $\mathsf{Tr}_G(e\bullet e')=\mathsf{Tr}_G(e)\circ \mathsf{Tr}_G(e')=A\circ B=A+B+AB$; then $AB\ne 0$ and consequently $f(A)f(B)=f(AB)\ne 0$. The equation
\begin{equation*}
    \begin{split}
       \mathsf{Tr}_H(c\bullet c') & = \mathsf{Tr}_H(c)\circ\mathsf{Tr}_H(c')\\
       & = f(A)\circ f(B)\\
       & =f(A)+f(B)+f(A)f(B)
    \end{split}
\end{equation*}
says that the matrix transformation corresponding to $c\bullet c'$ has three non-zero entries and so $c\bullet c'$ contains three paths. Then $c$ must be followed by $c'$ and this yields $u_l=u_{l'}$. Similarly, $u_k=u_{k'}$ can be shown. Therefore, $f$ gives a one-to-one mapping between the edges of $G$ and $H$.

To prove the correspondence between the nodes of two graphs, let $v_x$ be a node in $G$, connected to $v_i$ in which $j\ne x$ and $C$ and $f(C)$ be the matrix transformations of $a:v_i\rightarrow v_x\in\mathsf{Mod}(G)$ and $b:u_y\rightarrow u_z\in\mathsf{Mod}(H)$, respectively. Since $e'$  is followed by $a$ in $\mathsf{Mod}(G)$, with the same reasoning as above, $c'$ must be followed by $b$ in $\mathsf{Mod}(H)$ and this means $u_k=u_y$. So $f$ also gives a one-to-one mapping between nodes of graphs compatible with edges. Then, $G$ and $H$ are isomorphic.
 \end{proof}
 \subsection{Proof of Theorem \ref{cover of neighborhoods}}
\begin{figure}[h!]
    \centering
\includegraphics[scale=0.42]{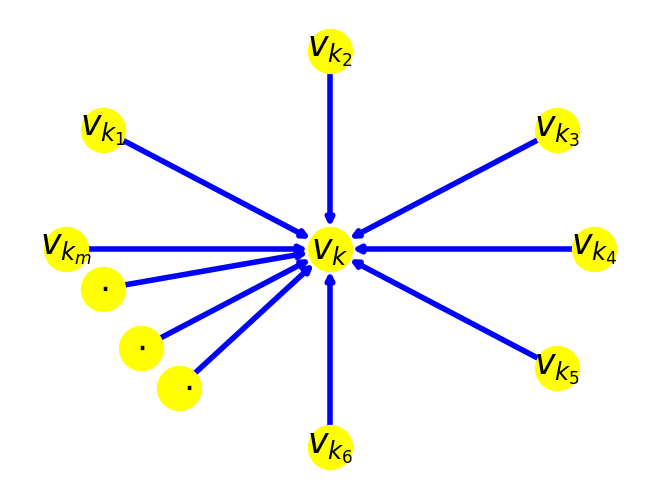}
    \caption{Visualizing a neighborhood by representing it as a directed subgraph}
    \label{sink}
\end{figure}

The role of neighborhoods in MPNN is like a sink such that messages move to the center of the sink. For a node $v_k$ with neighborhood $N_k$ containing $v_{k_1}, v_{k_2}, \cdots, v_{k_m}$, we depict this sink in Figure \ref{sink} by denoting directed edge from 
$v_{k_i}$ to $v_k$ by $e_i:v_{k_i}\rightarrow v_k$.
This sink can be considered as a directed subgraph. As an element of $\mathsf{Mod}(G)$, it can be represented as follows:
\[S_k=e_1\bullet e_2 \bullet\cdots\bullet e_m\]
Since the directed edges $e_i$ and $e_j$ appearing in $S_k$ are not composable, we observe $e_i\bullet e_j=e_j\bullet e_i$, rendering the order in $S_k$ unimportant. The cover obtained by $S_k$s is exactly the cover of the neighborhoods. Let $T_k=\mathsf{Tr}(S_k)$ and $A_i=\mathsf{Tr}(e_i)$. Thus $A_i$ has $1$ in the entry $k_i k$ and $0$ for all other entries.  The matrix transformation of $e_i\bullet e_j$ has just two non-zero entries and $\mathsf{Tr}(e_i\bullet e_j)=A_i+A_j+A_iA_j$. Then $A_iA_j=0$ for $1\le i\le m$ and $1\le j\le m$.
Theorem \ref{representation} implies
\begin{equation*}
    \begin{split}
    T_k=\mathsf{Tr}(S_k)&=A_1\circ A_2\circ \cdots \circ A_m\\
    &=A_1 + A_2 + \cdots + A_m
    \end{split}
    \end{equation*}
As a result, the column $k$ of $T_k$ aligns with the column $k$ of the adjacency matrix of graph $G$, while the remaining columns are filled with zeros. Transforming the cover $\lbrace S_k\rbrace$ yields a collection of $|V|$ matrices, each containing a single column from the adjacency matrix. In the GkGNN framework, summation is an allowed operation, enabling the construction of the adjacency matrix by performing the summation on this matrix collection. Hence, neighborhoods can function as a cover within the framework of GkGNN, with the adjacency matrix serving as an interpretation of this cover.
\subsection{Proof of Theorem \ref{coversieveinvariant}}
 \begin{proof}
     Since the definition of sets $M_i(v)$s is based on the neighborhoods, for a graph isomorphism $f: G\rightarrow H$, $f(M_i(v))=M_i(f(v))$. This follows $\mathsf{Mod}(f)(D_i(v))=D_i(f(v))$. Since $\mathsf{Mod}(f)$ is a monoidal homomorphism, we get:
\begin{center}
\small
    \begin{equation*}
        \begin{split}
            \mathsf{Mod}(f)(\mathsf{Sieve}(v,k)) & = \mathsf{Mod}(f)(D_k(v)\bullet \cdots\bullet D_0(v))\\
            & =\mathsf{Mod}(f)(D_k(v))\bullet\cdots\bullet\mathsf{Mod}(f)(D_0(v))\\
            & = D_k(f(v))\bullet\cdots \bullet D_0(f(v))\\
            & =\mathsf{Sieve}(f(v),k)
        \end{split}
    \end{equation*}
\end{center}
    Based on Theorem \ref{iso_graph-induce}, $\mathsf{Mom}(f)(\mathsf{Image}(v,k))=\mathsf{Image}(f(v),k)$.
 \end{proof}
 \subsection{Proof of Theorem \ref{snn invariant}}
 \begin{proof}
    Since the cover of sieves is invariant
    and $\mathsf{CO}(f)$ preserves the rest of the computations in the algorithm, $\mathsf{SNN}$ is invariant.
\end{proof}
\subsection{Proof of Theorem \ref{monoid adj}}
 \begin{proof}
     Let $\mathsf{Adj}(v)$ denote the matrix representation of the neighborhood of a node $v \in G$. As demonstrated, this matrix contains exactly one non-zero column. The mapping $f$ is a Change-of-Order mapping, which transforms $\mathsf{Adj}(v)$ into a matrix with a single non-zero column, where all non-zero entries are equal to $1$.

An element of $\mathsf{Adj}(H)$ that is not a matrix transformation of any element in the cover of the neighborhood will have two or more non-zero columns. Consequently, for $f(\mathsf{Adj}(v)) \in \mathsf{Adj}(H)$, there exists a node $u \in H$ such that $f(\mathsf{Adj}(v)) = \mathsf{Adj}(u)$.

This establishes a one-to-one correspondence between $V_G$ and $V_H$, as $f$ is an isomorphism.
Now, let $\xymatrix{v_i \ar@{-}[r] & v_j}$ represent an edge in $G$, with $f(\mathsf{Adj}(v_i)) = \mathsf{Adj}(u_k)$ and $f(\mathsf{Adj}(v_j)) = \mathsf{Adj}(u_l)$. The entry $ii$ in the matrix $\mathsf{Adj}(v_j) \circ \mathsf{Adj}(v_i)$ equals $1$.

Since $f$ is a Change-of-Order mapping, the matrix $f(\mathsf{Adj}(v_j) \circ \mathsf{Adj}(v_i)) = \mathsf{Adj}(u_l) \circ \mathsf{Adj}(u_k)$ has a diagonal entry equal to $1$. In this matrix, the only diagonal entry that can be non-zero is the entry $kk$. Similarly, the entry $ll$ in $\mathsf{Adj}(u_k) \circ \mathsf{Adj}(u_l)$ equals $1$. This implies that there is an edge between $u_k$ and $u_l$.

Thus, we establish a one-to-one correspondence between the edges of $G$ and $H$ that is consistent with the mapping of their nodes. This proves that $f$ defines a graph isomorphism between $G$ and $H$.
 \end{proof}
\end{document}